\pdfoutput=1

\documentclass[11pt]{article}

\usepackage[preprint]{acl}
\usepackage{times}
\usepackage{latexsym}

\usepackage[T1]{fontenc}

\usepackage[utf8]{inputenc}

\usepackage{microtype}

\usepackage{inconsolata}

\usepackage{tcolorbox}
\usepackage{amsthm}
\usepackage{multirow}
\usepackage{graphicx}
\usepackage{colortbl} 
\usepackage{booktabs}
\usepackage{bm}
\usepackage{amsfonts}
\usepackage{pgfplots}
\usepackage{amsmath}
\usepackage{utfsym}
\usepackage{soul}
\usepackage{color, xcolor} 
\usepackage{amssymb}
\usepackage[normalem]{ulem}
\usepackage{ulem}
\definecolor{colora}{HTML}{DAF1FE}
\definecolor{lightgray}{HTML}{F0F0F0}

\DeclareRobustCommand{\hlcolorb}[1]{{\sethlcolor{lightgray}\hl{#1}}}

\title{How Reliable are LLMs as Knowledge Bases? Re-thinking \\Facutality and Consistency}

\author{Danna Zheng$^{1}$, Mirella Lapata$^{1}$, Jeff Z. Pan$^{1, 2}$ \\
            $^{1}$ School of Informatics, University of Edinburgh, UK\\
            $^{2}$ Huawei  Edinburgh Research Centre, CSI, UK\\
        dzheng@ed.ac.uk, mlap@inf.ed.ac.uk, http://knowledge-representation.org/j.z.pan/
}

\begin{document}
\maketitle
\begin{abstract}
Large Language Models (LLMs) are increasingly explored as knowledge bases (KBs), yet current evaluation methods focus too narrowly on knowledge retention, overlooking other crucial criteria for reliable performance.
In this work, we rethink the requirements for evaluating reliable LLM-as-KB usage and highlight two essential factors: factuality, ensuring accurate responses to seen \emph{and} unseen knowledge\footnote{\textit{Seen knowledge} refers to knowledge learned during training. \textit{Unseen knowledge}  is neither  present in  the model's training data nor can be inferred from seen knowledge.}, and consistency, maintaining stable answers to questions about the same knowledge.
We introduce \texttt{UnseenQA}, a dataset designed to assess LLM performance on unseen knowledge, and propose new criteria and metrics to quantify factuality and consistency, leading to a final reliability score.
Our experiments on 26 LLMs reveal several challenges regarding their use as KBs,  underscoring the need for more principled and comprehensive evaluation.

\end{abstract}
\section{Introduction}
Large Language Models (LLMs), pretrained on vast text corpora, have demonstrated significant capabilities in encoding knowledge without explicit supervision. The continuous release of new LLMs, evaluated on benchmarks like TriviaQA~\cite{joshi2017triviaqa} and Natural Questions~\cite{kwiatkowski2019natural}, highlights their improving ability to answer fact-based queries. This progress has fueled interest in employing LLMs as knowledge bases (KBs) for various applications and developing techniques to edit model knowledge~\cite{wang-etal-2024-retrieval,wang-etal-2024-detoxifying,wang-etal-2024-roselora} or mitigate hallucinations~\cite{zhang-etal-2024-self,zhang-etal-2024-truthx,yu-etal-2024-truth}.

However, a critical question remains underexplored: \textit{What criteria should an LLM meet to function reliably as a KB?} Current research often assumes that knowledge retention alone is sufficient~\cite{sun2023head, wang2021can, roberts2020much}. Existing evaluations generally follow two approaches: (1) converting knowledge graphs into natural language questions and assessing how many questions the LLM answers correctly~\cite{petroni2019language, sun2023head}; and (2) pretraining LLMs on knowledge-rich text and measuring their accuracy on related questions~\cite{wang2021can, he2024can}.

\begin{figure*}[t]
    \centering
    \includegraphics[width=\textwidth]{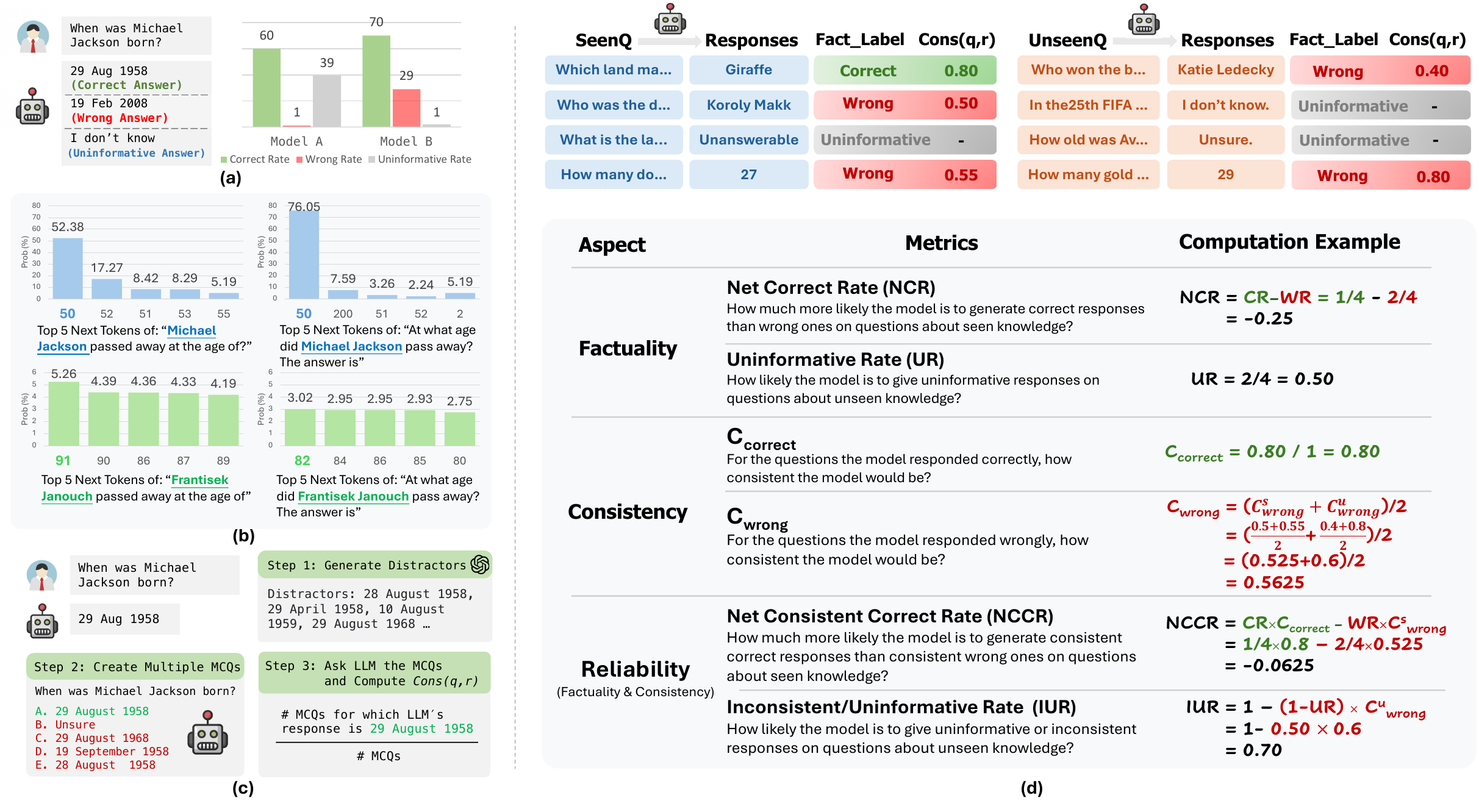}
    \caption{(a) An example illustrating three answer types: correct, wrong, and uninformative.  Focusing only on the correct rate  incorrectly suggests that Model B is better, even though  Model A is more reliable with a similar correct rate and a much lower wrong rate. (b) Illustration of LLM inconsistency with  \textsc{davinci-002} (temperature is set to~0). Questions in the top focus on seen knowledge, with  probability distribution mass concentrated on one prediction. Questions in the bottom  focus on unseen knowledge, where the distribution is more even. Drawing from such a distribution inevitably leads to inconsistencies. (c) Example computation for consistency score~$Cons(q, r)$. The  LLM's original answer is shown in \textcolor[HTML]{529E3E}{green}, while distractors are \textcolor{red}{red}. (d) An example illustrating how to evaluate LLM-as-KB. } 
    \label{fig:illustration}
\end{figure*}

These methods demonstrate that LLMs can recall information, but knowledge volume alone does not guarantee reliable performance as a KB. Beyond retention, it is essential to examine how LLMs handle factual queries—specifically, whether they respond accurately to seen knowledge and avoid making claims about unseen knowledge (\textit{factuality}), and whether they provide consistent answers to questions about the same facts (\textit{consistency}).

\href{https://www.vocabulary.com/dictionary/factuality#:~:text=Definitions%20of%20factuality,synonyms%3A%20factualness}{Factuality} refers to the quality of being factual or based on fact. 
KBs hosted on servers or cloud platforms, offer precise answers or null responses when data is unavailable. In contrast, LLMs rely on probabilistic next-token prediction, which can lead to plausible but incorrect answers. As a result, LLM responses are typically correct, uninformative, or wrong. Existing methods for evaluating factuality often focus on the rate of correct answers in factual QA datasets~\cite{chen2023felm, wang2024factuality}. However, as many studies~\cite{lin2022truthfulqa, sun2023head} fail to specify whether the dataset’s knowledge was included in the LLM's pre-training data, it is not possible to establish whether the model is genuinely factual. 
Secondly, it is misleading to equate a higher correct rate with greater factuality. As illustrated in Figure~\ref{fig:illustration}(a), a model with a higher correct rate might still be less factual if it produces a higher rate of wrong answers compared to a model with fewer errors overall.

\href{https://www.oxfordlearnersdictionaries.com/definition/english/consistency}{Consistency} refers to the quality of always behaving in the same way or having the same opinions. 
Traditional KBs achieve consistency through algorithms~\cite{andersen2001easy} that detect and resolve conflicts. In contrast, LLMs frequently exhibit inconsistent behavior~\cite{elazar2021measuring, wang2022self}. 
Current research~\cite{elazar2021measuring, jang2022becel, hagstrom2023effect} evaluates LLM consistency using benchmarks involving paraphrasing, negation, or multilingual variations, favoring models that maintain consistent responses across diverse samples. 
However, we argue that expecting LLMs to always be consistent in fact-based responses is overly rigid. Unlike KBs, which store information in fixed locations, LLMs operate probabilistically. When the context has been learned during training, the probability distribution for predictions is concentrated; otherwise, it remains more uniform. Sampling from a uniform distribution naturally leads to inconsistencies. As shown in Figure~\ref{fig:illustration}(b), even with greedy decoding, slight distribution biases can cause variations in the top-selected words.

Given these issues, this paper seeks to define the criteria for a reliable LLM-as-KB when handling factual queries. In evaluating factuality, we consider both seen knowledge (contained within training data) and unseen knowledge, and consider the negative effects of wrong answers.
To assess performance on unseen knowledge, we introduce \texttt{UnseenQA}, a new dataset containing knowledge unavailable to LLMs trained before April 13, 2024. For consistency, we classify the correctness of responses and propose a novel method to compute the probability that an LLM can consistently provide the same response~$r$ to a question~$q$.

We evaluate 26 popular LLMs and find that: 
1) \textsc{GPT-3.5-Turbo} achieves balanced performance in both factuality and consistency. LLMs like \textsc{LLaMA3-70B} may achieve a higher correct rate on seen knowledge but exhibit higher wrong rates and consistency on wrong answers. 
2) There is a correlation between factuality and consistency: more factual LLMs tend to be consistent in their responses, whether correct or wrong. 
3) Larger LLMs perform worse on unseen knowledge and are more consistent even when providing wrong answers. 
4) Fine-tuning techniques can improve performance on unseen knowledge. However, this often comes at the expense of performance on seen knowledge.
5) In-context learning (ICL) does not improve performance on seen knowledge, as it as it generally increases or decreases both correct and wrong rates simultaneously.
6) Base LLMs tend to overestimate their knowledge on numerical and temporal questions.

We hope our work will draw the community’s attention to the multifaceted challenges of using LLMs-as-KBs, ultimately inspiring further research and innovation toward more reliable, robust, and principled methodologies.

\section{What is a Reliable LLM-as-KB? }

In simple terms, an LLM  is a reliable KB if it consistently provides factual responses.
As illustrating in Figure~\ref{fig:illustration} (d), evaluating the reliability of LLMs as KBs primarily involves assessing two critical dimensions, namely factuality and consistency.

\subsection{Factuality}

We propose the following criteria for determining the factuality of LLMs-as-KBs:
\begin{tcolorbox}
    \textbf{Criterion 1.1}: For seen knowledge,  a factual LLM should demonstrate a high correct rate and a low wrong rate.

    \textbf{Criterion 1.2}: For unseen knowledge, a factual LLM should demonstrate a high uninformative rate.   
\end{tcolorbox}

We next proceed to define evaluation metrics that operationalize  these criteria.
Let $\boldsymbol{M}$ denote an LLM.  Let $D_{\text{seen}}$ denote a QA dataset containing $N$~open-ended factoid questions pertaining to  knowledge the LLM ought to have seen during training. Let~$D_{\text{unseen}}$ denote a QA dataset with~$L$ open-ended factoid questions covering  unseen knowledge. 
We further assume the LLM's response to $D_{\text{seen}}$ will be correct, uninformative, or wrong, while its response to $D_{\text{unseen}}$ will be either  uninformative or wrong. 

\paragraph{\hlcolorb{METRIC 1.1: Net Correct Rate (NCR)}}\hspace*{-1.2ex}measures how much more likely the model is to provide correct responses instead of wrong ones on $D_{\text{seen}}$~questions. It is defined as:
\begin{equation}
    \text{NCR} = \text{CR} - \text{WR}
\end{equation}
\begin{align}
\text{CR}=\frac{N_{correct}}{N} & &
\text{WR}=\frac{N_{wrong}}{N}
\end{align}
where $N_{correct}$ and $N_{wrong}$ are  counts of correct and wrong responses, respectively.

NCR values range from~$-1$ to~$1$. A negative NCR suggests the model tends to  provide misleading responses, while a positive NCR suggests a preference for correct responses.  Consider again two models, A and B. According to Criterion~1.1, if model A has a higher correct rate and lower wrong rate compared to model B, then model A is better. 
Formally, if $\text{CR}_A - \text{CR}_B > \text{WR}_A - \text{WR}_B$, then model A is better than B. Algebraically, this is equivalent to $\text{CR}_A - \text{WR}_A > \text{CR}_B - \text{WR}_B$, i.e.,~$\text{NCR}_{A} > \text{NCR}_{B}$. Therefore, a higher NCR indicates a more factual model on seen knowledge.

\paragraph{\hlcolorb{METRIC 1.2: Uninformative Rate (UR)}}\hspace*{-1.2ex}assesses whether the model is likely to provide  uninformative responses to  $D_{\text{unseen}}$ questions. It is formulated as:
\vspace{-.2cm}
\begin{equation}
\text{UR}=\frac{L_{uninformative}}{L}
\end{equation}
where $L_{uninformative}$ denotes the count of uninformative responses. 
UR ranges from 0 to 1. A higher UR indicates that the model is more likely to refrain from giving wrong responses when faced with unseen knowledge.

\subsection{Consistency}

We propose the following  consistency criteria:
\begin{tcolorbox}
    \textbf{Criterion 2.1}: The model is expected to be consistent in correct responses.

    \textbf{Criterion 2.2}: The model is expected to be inconsistent in wrong responses.
\end{tcolorbox}

We next define evaluation metrics corresponding to the criteria above. 
Let $q$~refer to a question in either $D_{\text{seen}}$ or $D_{\text{unseen}}$, and $r$ denote model $M$'s response to $q$.
Inspired by \citet{zheng2024trustscore}, we measure consistency  based on multiple-choice questions (MCQs).
As shown in Figure~\ref{fig:illustration} (c), we employ \textsc{gpt-3.5-turbo-insruct} to generate a set of distractor options similar to response~$r$, and then create a group of MCQs. The consistency score for data point $(q, r)$ is calculated as:

\begin{equation}
\text{Cons}(q, r) = \frac{\sum_{i=1}^{X_{\text{MCQs}}} [R_i = r]}{X_{\text{MCQs}}}
\end{equation}
where $X_{\text{MCQs}}$ is the total number of MCQs, $R_i$~is model $M$'s response for the $i$-th MCQ, and~$[R_i=r]$ yields 1 when the model's response $R_i$ matches its original response $r$, and 0~otherwise.
The consistency score $\text{Cons}(q, r)$ ranges from 0 to 1.

\paragraph{\hlcolorb{METRIC 2.1: $C_{\textit{correct}}$}}\hspace*{-1.2ex}measures a model's consistency in its correct responses and is defined as:
\begin{equation}
    C_{\textit{correct}} = \frac{\sum_{j=1}^{N_{correct}} Cons(q_j^{(c)}, r_j ^{(c)})}{N_{correct}}
\end{equation}
where $r^{(c)}$ refers to the response labeled as correct, and $q^{(c)}$ is the corresponding question.
$C_{\textit{correct}}$ ranges from~0 to 1. Based on Criterion~2.1, a higher $C_{\textit{correct}}$ is desirable.

\paragraph{\hlcolorb{METRIC 2.2: $C_{\textit{wrong}}$}}\hspace*{-1.2ex}measures the consistency of an LLM when it provides wrong responses and is defined as:
\begin{equation}
    C_{\textit{wrong}} = \frac{C_{\textit{wrong}}^{s}+C_{\textit{wrong}}^{u}}{2}
\end{equation}
where $C_{\textit{wrong}}^{s}$/$C_{\textit{wrong}}^{u}$  refer to the consistency of an  LLM when it provides wrong responses to questions about seen/unseen knowledge:
\begin{equation}
    C_{\textit{wrong}}^{s} = \frac{\sum_{j=1}^{N_{wrong}} Cons(q_j^{(w)}, r_j ^{(w)})}{N_{wrong}}
\end{equation}
\begin{equation}
    C_{\textit{wrong}}^{u} = \frac{\sum_{j=1}^{L_{wrong}} Cons(q_j^{(w)}, r_j^{(w)})}{L_{wrong}}
\end{equation}
where $r^{(w)}$ and $q^{(w)}$ denote wrong responses and their corresponding questions. $N_{wrong}$ and $L_{wrong}$ are the counts of wrong answers on $D_{\text{seen}}$ and $D_{\text{unseen}}$, respectively. $C_{\textit{wrong}}$ ranges from 0 to 1, and per Criterion~2.2, lower $C_{\textit{wrong}}$ is better.

\subsection{Reliability (Factuality and Consistency)}
Based on the criteria defined above, an LLM is reliable as a KB if it meets the following criteria when evaluated against factuality \emph{and} consistency:
\begin{tcolorbox}
    \textbf{Criterion 3.1}: For seen knowledge, an LLM should have a high rate of consistently correct responses and a low rate of consistently wrong responses.

    \textbf{Criterion 3.2}: For unseen knowledge, a  LLM  should have a high rate of uninformative or inconsistent responses.
\end{tcolorbox}
We next quantify these criteria are with the following two metrics.

\paragraph{\hlcolorb{METRIC 3.1: Net Consistently Correct Rate (NCCR)}}\hspace*{-1.2ex}quantifies the model's tendency to provide consistently correct responses compared to consistently wrong ones for questions about seen knowledge. It is defined as:
\begin{align}
    \text{NCCR} & =   \text{CCR} - \text{CWR}\\\nonumber
    \text{CCR} & = \text{CR} \times C_{correct} \\\nonumber
\text{CWR} & =  \text{WR} \times C_{wrong}^{s}
\end{align}
NCCR ranges from $-1$ to $1$.  NCCR values closer to~1 indicate an LLM is more reliable on seen knowledge.  
A negative NCCR  suggests the model provides consistently wrong responses, while a positive NCCR suggests a preference for consistently correct responses. 

\paragraph{\hlcolorb{METRIC 3.2: Inconsistent/Uninformative Rate (IUR)}}\hspace*{-1.2ex}assesses whether an LLM is likely to provide  uninformative or inconsistent wrong responses for questions about unseen knowledge. It is defined as:
\begin{equation}
    \text{IUR} = 1 - (1-\text{UR})C_{wrong}^{u}
\end{equation}
IUR ranges from 0 to 1. A higher IUR value indicates the LLM functions as a more reliable KB on unseen knowledge.
\section{Experimental Setup}
\subsection{LLM Selection}
We evaluate 26 popular LLMs, including \href{https://platform.openai.com/docs/models/gpt-3-5-turbo}{\textsc{gpt-3.5-turbo}}, \href{https://arxiv.org/abs/2210.11416}{\textsc{Flan-t5}}, \textsc{\href{https://arxiv.org/abs/2302.13971}{llama1}}, \textsc{\href{https://arxiv.org/abs/2307.09288}{llama2}}, \textsc{\href{https://arxiv.org/abs/2302.13971}{llama3}}, 
\textsc{\href{https://arxiv.org/abs/2310.06825}{mistral}}, 
\textsc{\href{https://arxiv.org/abs/2403.08295}{gemma}}, and 
\textsc{\href{https://www.microsoft.com/en-us/research/blog/phi-2-the-surprising-power-of-small-language-models/}{Phi2}}. 
Detailed descriptions of the evaluated LLMs are provided in Table~\ref{tab:llms-detail} in Appendix~\ref{sec:App-llm}. Detailed descriptions are available in Table~\ref{tab:llms-detail} in Appendix~\ref{sec:App-llm}. We test LLMs of different sizes: small (0.08B–3B), medium (7B–13B), and large (65B–70B). "Fine-tuned LLMs" refer to those fine-tuned via instruction-tuning or reinforcement learning from human feedback (e.g., \textsc{llama3instruct-8B}), while "base LLMs" refer to models without fine-tuning (e.g., \textsc{llama3-8B}).

\begin{table*}[h]
\centering
\tiny
\resizebox{\textwidth}{!}{%
\begin{tabular}{ccl}
\hline
Answer Type & Abb & \multicolumn{1}{c}{Template} \\ \hline
\multirow{4}{*}{Number} & T1 & How many gold medals did {[}country/region{]} win at the XXXIV Summer Olympic Games? \\
 & T2 & In the 25th FIFA World Cup, what was the final ranking of {[}country/region{]}? \\
 & T3 & How many children does {[}person{]} have? \\
 & T4 & How old was {[}person{]} in 2015? \\ \hline
\multirow{4}{*}{Person} & T5 & Who won the bronze medal of {[}medal event{]} at the XXXIII Summer Olympic Games? \\
 & T6 & Who is the supreme leader of {[}country/region{]} in 2040? \\
 & T7 & In 2028, who served as the head coach of {[}country/region{]} national football team? \\
 & T8 & Who is {[}person{]}'s mom? \\ \hline
\multirow{4}{*}{Time} & T9 & On which date was {[}person{]} born? \\
 & T10 & In what year did {[}person{]} die? \\
 & T11 & In what year did {[}person{]} graduate with the bachelor's degree? \\
 & T12 & When was the wedding date for {[}person{]}? \\ \hline
\multirow{4}{*}{Location} & T13 & Where was {[}person{]} born? \\
 & T14 & Where did {[}person{]} pass away? \\
 & T15 & Which university did {[}person{]} attend for the undergraduate studies? \\
 & T16 & Where was {[}person{]}'s wedding held? \\ \hline
\multirow{4}{*}{Others} & T17 & What was the cause of {[}person{]}'s death? \\
 & T18 & What is the title of the debut album released by {[}person{]}? \\
 & T19 & What is the name of the first film directed by {[}person{]} \\
 & T20 & What is the occupation of {[}person{]}? \\ \hline
\end{tabular}%
}
\caption{Question templates used to create UnseenQA}
\label{tab: temps}
\end{table*}

\subsection{Datasets}

\paragraph{SeenQA} 
\texttt{SeenQA} is a composite dataset comprising 3,000 questions sourced from the test sets (or development sets, where test sets were unavailable) of \href{https://huggingface.co/datasets/vasudevgupta/natural-questions-validation}{Natural Questions}, \href{https://huggingface.co/datasets/mandarjoshi/trivia_qa/viewer/rc.wikipedia.nocontext}{TriviaQA}, and \href{https://huggingface.co/datasets/akariasai/PopQA}{PopQA} (see Appendix~\ref{sec:App-data}).
All three datasets are derived from Wikipedia, with a knowledge cutoff date no later than December 2018. Given that Wikipedia is a frequent source for pre-training LLMs and the evaluated LLMs in this study have a knowledge cutoff date beyond April 2019 (as shown in Table~\ref{tab:llms-detail}), it can be inferred that the knowledge in these datasets is available to  LLMs during their training. 
The creation of \texttt{SeenQA} involved a three-step process:
1) \textbf{Factoid Question Extraction}: exclude "why" questions, those with multiple answers, or answers exceeding five tokens.
2) \textbf{Time-Sensitive Question Removal}: use \textsc{gpt-4-1106-preview} (prompt in Table~\ref{tab:prompt_time_sen}) to detect and remove questions with time-variant answers.
3)~\textbf{Random Sampling}: randomly select 1,000 questions per source, discarding supporting context for a closed-book setting.

\paragraph{UnseenQA}
UnseenQA is a new QA dataset designed to ensure the LLMs tested in this study lack access to knowledge required to answer questions. It consists of 3,000 questions generated from 20 hand-written templates (150 questions per template), as detailed in Table~\ref{tab: temps}, spanning five answer types: number, person, time, location, and others.
Templates T1--T7 focus on future events with unknown answers at the time of writing, while the remaining templates involve fictional individuals whose names and details are not available online. The templates feature three placeholder types:
1) \textbf{Country/Region}: 150 names sourced from the National Olympic Committees listed on the~\href{https://en.wikipedia.org/wiki/2024_Summer_Olympics}{Wikipedia page}.
2) \textbf{Medal Event}: 150 medal events from the~\href{https://stillmed.olympics.com/media/Documents/Olympic-Games/Paris-2024/Paris-2024-Event-Programme.pdf}{official programme of the Olympic games, Paris 2024}.
3) \textbf{Person}: 150 randomly generated names from combinations of 100 first, middle, and last names, manually verified to lack online presence.
The dataset was created on April 13, 2024. 
All LLMs studied were released before this date and thus lacked access to the knowledge.

\subsection{Evaluation on a Single Response} 
\label{sec:eval_on_a_single_response}
\paragraph{Uninformative} We classify uninformative responses into three categories: \textbf{repetition}, \textbf{none}, and \textbf{unsure} (see Appendix~\ref{sec:App-unin-EX}).
\textbf{Repetition} refers to responses that repeatedly echo a specific string.
We detect this using regular expressions and word frequency analysis.
\textbf{None} includes responses that lack relevant content, such as empty strings or those merely repeating the question.
\textbf{Unsure} applies to responses where the model says it is unable to answer or does not know. Examples include phrases like "I am not sure" or "I am just an AI", etc.

\paragraph{Correct} A response is considered correct if it is an exact match with the ground truth or is similar enough as judged by  \textsc{\href{https://platform.openai.com/docs/models/gpt-4o}{GPT-4o}}.
Following prior work~\cite{sun2023head}, which reported a 98\% agreement rate between ChatGPT and human evaluations, we adopt their evaluation prompt (detailed in Table~\ref{tab:correct_check} in Appendix~\ref{sec:App-prompt}).

\begin{figure*}[t]
    \centering
    \includegraphics[width=\textwidth]{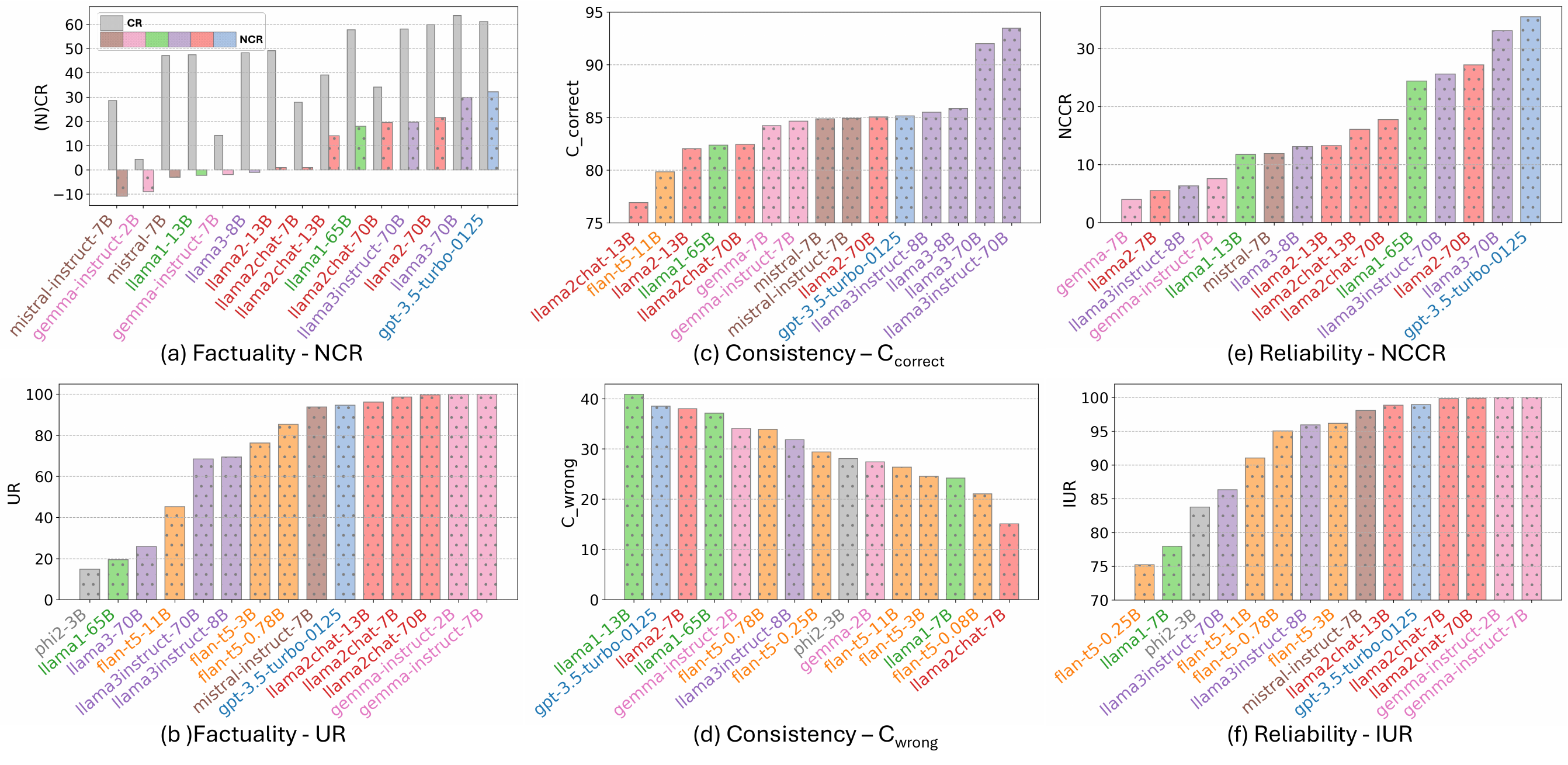}
    \caption{Top 15 LLMs ranked by (a) Net Correct Rate, (b) Uninformative Rate, (c) $C_{\textit{correct}}$, (d) $C_{\textit{wrong}}$, (d) Net Consistent Correct Rate, (e) Inconsistent/Uninformative Rate. Values are scaled by 100 (full results in Appendix~\ref{sec:App-ex}).} 
    \label{fig:result_mini}
\end{figure*}

\paragraph{Consistency Score} To compute $\text{Cons}(q, r)$,  we set $X_{\text{MCQs}}$ (total number of MCQs) to 20, and each MCQ includes question $q$ and 5 options (the original response $r$, 3 random distractor options, and an `unsure' option).

\subsection {Prompts and Hyper-parameters}
All LLMs were evaluated using greedy decoding (temperature 0 for \textsc{gpt-3.5-turbo}) with a maximum of 100 new tokens.
To provide a comprehensive evaluation, we experimented with three types of prompt settings: zero-shot, four-shot, and four-shot with two unsure shots.  
To avoid any bias introduced by fixed examples, we employed a dynamic few-shot method following the work of \citet{nori2023can}. For further details of the prompt settings, refer to Appendix~\ref{sec:App-prompt}.
\section{Quantitative Results}
We present results for all LLMs under different prompt settings  in 
in Appendix~\ref{sec:App-ex} (Table~\ref{tab:factulity}: factuality, Table~\ref{tab:consistency}: consistency, and Table~\ref{tab:reliability}: reliability). 
LLM rankings based on different metrics are shown in Figure~\ref{fig:rank} and Figure~\ref{fig:fcr}, and also  in Appendix~\ref{sec:App-ex}. In the remainder, we discuss our results in dynamic four-shot settings.

\paragraph{Which LLM is  most factual?} 
As shown in Figures~\ref{fig:result_mini} (a) and (b), \textsc{gpt-3.5-turbo} performs best on seen knowledge, while \textsc{gemma-instruct-7B} excels at unseen knowledge. \textsc{gpt-3.5-turbo} and \textsc{llama2chat-70b} emerge as the most balanced models, demonstrating top performance across both seen and unseen knowledge.
Notably, Figure~\ref{fig:result_mini} (a) highlights a limitation of CR, the standard metric for factuality. For instance, while \textsc{llama1-13b} has a significantly higher CR than \textsc{gemma-instruct-7B}, their behavior differs in terms of wrong responses. \textsc{llama1-13b} exhibits a 33.50\% higher WR rate, leading to a lower NCR compared to \textsc{gemma-instruct-7B}.

\paragraph{Which LLM is most consistent?} 
As shown in Figures~\ref{fig:result_mini} (c) and (d), \textsc{llama3instruct-70b} is most consistent on questions it answers correctly, while \textsc{llama2chat-7b} is  least consistent on questions it answers incorrectly. Among these models, \textsc{llama3instruct-8B} is most balanced, with a high $C_{\textit{correct}}$ and low $C_{\textit{wrong}}$.

\paragraph{Which LLM is most reliable?}
As shown in Figures~\ref{fig:result_mini} (e) and (f), \textsc{gpt-3.5-turbo} excels at seen knowledge, while \textsc{gemma-instruct-7B} performs best on unseen knowledge. Among these models, \textsc{gpt-3.5-turbo} demonstrates balanced performance, with a top NCCR and a high IUR. Note that a more factual LLM is not necessarily more reliable. For example, \textsc{llama3-instruct-8B} ranks below the 15th in NCR and 13th in NCCR due to its relatively high $C_{\textit{correct}}$ and low $C_{\textit{wrong}}$.

\section{Discussion}

\subsection{Correlation Analysis}
\begin{table}[t]
\centering
\resizebox{.48\textwidth}{!}{%
\begin{tabular}{@{}l@{~}c@{~}c@{~}c@{}}
\toprule
\textbf{Comparisons} & \multicolumn{1}{c}{\textbf{Zero-Shot}} & \multicolumn{1}{c}{\textbf{Four-Shot}} & \multicolumn{1}{c}{\textbf{Four-Two}} \\ \midrule

NCR vs. UR & 0.27 & 0.34 & \underline{0.62}  \\
NCCR vs. IUR  & \hspace*{-1.9ex}$-$0.17 &  \hspace*{-1.9ex}$-$0.12 & \underline{0.41}  \\
$C_{\textit{correct}}$ vs. $C_{\textit{wrong}}$ & \underline{0.81} & \underline{0.78} & \underline{0.51} \\ 
NCR vs. $C_{\textit{correct}}$ & \underline{0.60} & \underline{0.41} & 0.37 \\ 
NCR vs. $C_{\textit{wrong}}$ & \underline{0.48} & \underline{0.64} & \underline{0.41} \\ 
UR vs. $C_{\textit{correct}}$ & \underline{0.43} & 0.07 & 0.35 \\
UR vs. $C_{\textit{wrong}}$ & \underline{0.34} & 0.05 & \underline{0.48} \\
\bottomrule
\end{tabular}%
}
\caption{Pairwise correlation of LLM performance on different metrics under different prompt settings.  The correlations are computed across all LLMs (26~data points). We report Pearson's $\rho$, with underlined values indicating statistical significance ($p<0.05$). Four-Two refers to the four-shot setting with two unsure shots.
}
\label{tab:PRs}
\end{table}

\paragraph{Does an LLM with strong performance on seen knowledge tend to perform strong on unseen knowledge?}
We next examine whether there is a linear relationship between performance on seen and unseen knowledge.
Performance on seen knowledge does not reliably predict performance on unseen knowledge in zero-shot and few-shot settings (without unsure shots).
Table~\ref{tab:PRs} (the first and second rows) shows the correlation values (Pearson's~$\rho$) between NCR and UR, and NCCR and IUR, revealing no statistically significant correlations in these settings.
However, in the four-shot setting with two unsure shots, correlations across all metrics are significant. While seen knowledge performance does not transfer to unseen knowledge, specific prompt manipulations can improve these correlations (see the last column in Table~\ref{tab:PRs}).

\paragraph{Does an LLM with high consistency score on correctly answered questions tend to perform less consistent on wrongly answered questions?} 
As shown in Figure~\ref{fig:C_correct_and_C_wrong} and Table~\ref{tab:consistency} in Appendix~\ref{sec:App-ex}, an LLM tends to be more consistent on questions it answers correctly than on those it answers wrongly. 
However, when comparing different LLMs, models with higher $C_{\textit{correct}}$ also tend to have higher $C_{\textit{wrong}}$, as indicated by the positive, significant correlation reported in Table~\ref{tab:PRs} (the third row).
This finding suggests that LLMs consistent with correct responses are also consistent with wrong responses, contradicting the expectation of high $C_{\textit{correct}}$ and low $C_{\textit{wrong}}$. This highlights a notable flaw in current models, which future work should address.

\paragraph{Does strong factuality performance correlate with strong consistency performance?}
As shown in Table~\ref{tab:PRs} (last four rows), there is a positive correlation between NCR/UR and $C_{\textit{correct}}$/$C_{\textit{wrong}}$, particularly in zero-shot settings. This suggests that LLMs with strong factuality performance tend to be confident in their responses, maintaining consistency whether those responses are correct or wrong.

\subsection{Model Size, Fine-tuning and ICL Impact}
\begin{figure}[t]
    \centering
    \includegraphics[width=.5\textwidth]{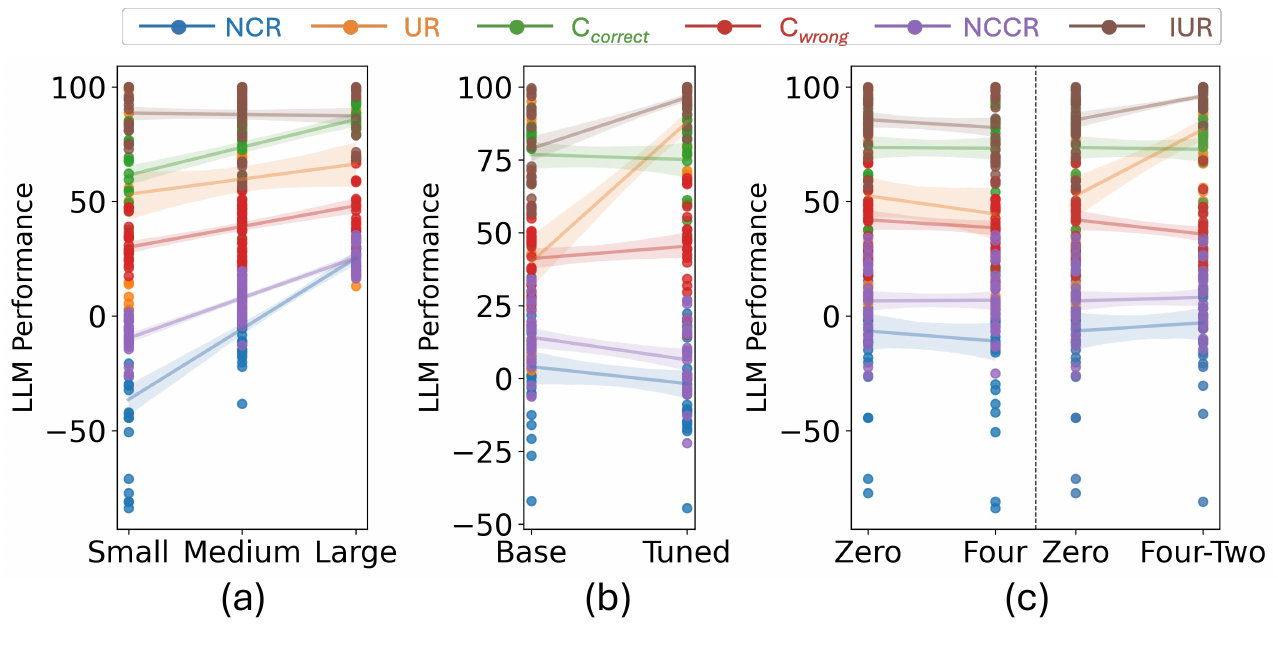}
    \caption{The impact of (a) model size, (b) fine-tuning, (c) ICL on LLM performance, measured with NCR, UR, C$_{correct}$, C$_{wrong}$, NCCR, and IUR. Different metrics are color-coded. See Appendix~\ref{sec:App-ex} for more detailed visualization. Values are scaled by 100.} 
    \label{fig:size_ft_icl}
\end{figure}

\begin{figure*}[t]
    \centering
    \includegraphics[width=\textwidth]{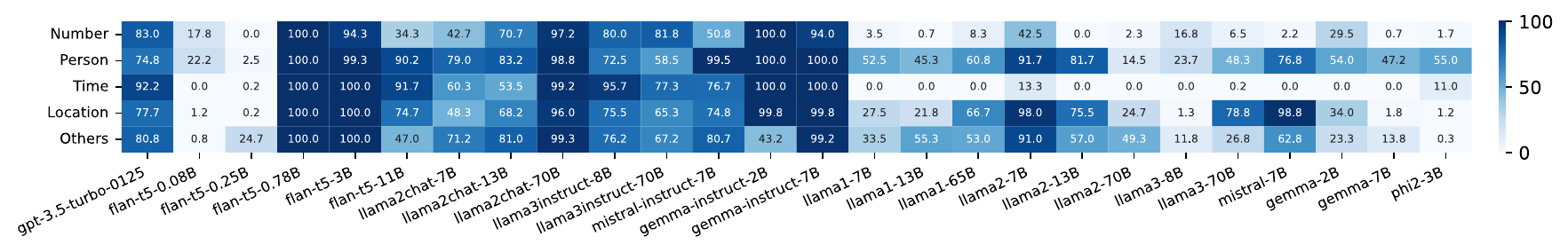}
    \caption{The impact of question type on LLM performance, measured by Uninformative Rate (UR) on unseen knowledge (scaled by 100). Questions are grouped by answer type. Higher values have darker shades.}
    \label{fig:question-type}
\end{figure*}
\paragraph{How does model size affect LLMs performance?} 
As shown in Figure~\ref{fig:size_ft_icl}(a), \textbf{\textit{larger LLMs perform better on seen knowledge but worse on unseen knowledge}}.
As model size increases, both NCR (\textcolor[HTML]{3D76AF}{blue line}) and NCCR (\textcolor[HTML]{8D69B7}{purple line}) improve, indicating better performance on questions related to seen knowledge. 
However, while the UR (\textcolor[HTML]{EB8635}{orange line}) increases, the IUR (\textcolor[HTML]{84584D}{brown line}) decreases. 
This suggests that larger models make fewer wrong responses on unseen knowledge but tend to give more consistent wrong responses.
We also observe that \textbf{\textit{larger LLMs are more consistent, even with wrong responses}}. Both $C_{\textit{correct}}$ (\textcolor[HTML]{529E3E}{green line}) and $C_{\textit{wrong}}$ (\textcolor[HTML]{C53932}{red line}) rise significantly with model size. While higher consistency in correct responses is desirable, the increase in consistency for wrong responses presents a risk. Larger models may confidently and consistently produce incorrect yet convincing information, heightening the potential for misinformation if not carefully managed.

\paragraph{How does fine-tuning affect LLMs performance?} 
Figure~\ref{fig:size_ft_icl}(b) suggests that \textbf{\textit{fine-tuning improves performance on unseen knowledge but degrades performance on seen knowledge}}.
After fine-tuning, both UR (\textcolor[HTML]{EB8635}{orange line}) and IUR (\textcolor[HTML]{84584D}{brown line}) increase significantly, indicating improved handling of unseen knowledge. However, the decline in NCR (\textcolor[HTML]{3D76AF}{blue line}) and  NCCR (\textcolor[HTML]{8D69B7}{purple line})  shows that fine-tuning reduces the model's effectiveness with seen knowledge.
We also see that \textbf{\textit{fine-tuning has no impact on consistency}}. 
There is no notable change in $C_{\textit{wrong}}$ (\textcolor[HTML]{C53932}{red line}) or $C_{\textit{correct}}$ (\textcolor[HTML]{529E3E}{green line}) after fine-tuning which suggests that current instruction-tuning and RLHF methods do not improve LLM consistency.

\paragraph{How does ICL affect LLMs performance?} 
Figure~\ref{fig:size_ft_icl}(c) shows that \textbf{\textit{ICL does not improve performance on seen knowledge, but unsure shots enhance performance on unseen knowledge}}. 
ICL does not increase NCR (\textcolor[HTML]{3D76AF}{blue line}) or NCCR (\textcolor[HTML]{8D69B7}{purple line}), which contrasts with previous findings on LLMs’ ICL ability~\cite{brown2020language,chada-natarajan-2021-fewshotqa,touvron2023llama,bai2023qwen}. Prior research largely focuses on the \emph{correct rate} in few-shot settings without unsure shots. As shown in Table~\ref{tab:factulity} and Figure~\ref{fig:CRNR} (Appendix~\ref{sec:App-ex}), while the four-shot setting significantly increases the correct rate, it also raises the wrong rate. Adding unsure shots helps reduce the wrong rate but also lowers the correct rate. 
For unseen knowledge, incorporating two unsure shots in the four-shot setting substantially improves UR (\textcolor[HTML]{EB8635}{orange line}) and IUR (\textcolor[HTML]{84584D}{brown line}), indicating better handling of unknown questions. \textbf{\textit{ICL with unsure shots reduces consistency in wrong responses.}} Adding two unsure shots in the four-shot setting decreases $C_{\textit{wrong}}$ (\textcolor[HTML]{C53932}{red line}). Without unsure shots, there is no significant change in $C_{\textit{wrong}}$ (\textcolor[HTML]{C53932}{red line}) or $C_{\textit{correct}}$ (\textcolor[HTML]{529E3E}{green line}).

\subsection{Behavior on Unseen Knowledge}
\paragraph{How do LLMs perform on different types of questions about unseen knowledge?}
Figure~\ref{fig:question-type} shows that base LLMs tend to overestimate their knowledge when answering numerical and temporal questions. Their UR scores are significantly lower for these types of queries, indicating a tendency to provide misleading responses even when they lack  relevant knowledge.

\section{Related Work}

\citet{petroni2019language} first explored using pre-trained LMs as KBs, introducing LAMA and showing that BERT retains relational knowledge with precision as the metric.
\citet{roberts2020much} evaluated knowledge storage and retrieval using natural language queries, measuring accuracy.
\citet{wang2021can} fine-tuned BART with related passages to instill factual knowledge, assessing masked span recovery accuracy.
\citet{he2024can} trained T5 and LLaMA2 on Wikidata to evaluate large-scale knowledge retention via exact match and F1 scores.
\citet{sun2023head} tested LLMs on 18,000 fact-based QA pairs, reporting both accuracy and hallucination rates.

Many studies have highlighted the issue of factuality in current LLMs.
To address this, new benchmarks have been proposed to assess factuality~\cite{muhlgay-etal-2024-generating,zhao2024felm,liu2024evaluating}, although these still rely on correct rate as the primary metric.
Several methods to enhance LLM factuality have also been introduced~\cite{wang2022self,hase2024does,cohen2024evaluating,qin2024does}.
However, evaluating these improved models falls outside the scope of our experiments and is suggested for future work.

Research on consistency~\cite{rajan-etal-2024-knowledge,sreekar-etal-2024-axcel,saxena2024evaluating} has shown that LLMs often struggle with providing consistent responses. These works, however, do not differentiate between the consistency expectations for correctly and wrongly answered questions.

Compared to previous research, we propose a comprehensive framework to evaluate not only whether LLMs recall \emph{seen} knowledge but also their ability to respond to \emph{unseen} knowledge. In addition, we evaluate LLM consistency when answering questions about \emph{identical} knowledge.

\section{Conclusion}
In this paper, we rethink the requirements for evaluating LLMs as KBs and propose criteria emphasizing  factuality and consistency, and the combination thereof which we argue is an indicator of reliability. 
We proposed various metrics operationalizing these criteria and used them to assess LLM performance when answering questions pertaining to both seen and unseen knowledge.
We evaluated 26 LLMs on our newly proposed \texttt{SeenQA} and \texttt{UnseenQA} datasets, and examined the impact of model size, fine-tuning, and ICL.
Our experimental results highlight  the critical need for continued research to develop more robust strategies that ensure both factuality and consistency, enabling LLMs to reliably function as KBs.
\newpage
\section*{Limitations}
First, due to budget constraints, we conducted in-depth evaluations on only a single closed-source LLM (\textsc{GPT-3.5-turbo}). Nevertheless, this limitation does not diminish the contributions of our work. Our study includes a broad comparative analysis of 26 different LLMs, ensuring that the insights gained are comprehensive and not confined to any single model. Furthermore, the primary objective of our paper is to introduce a systematic framework for evaluating LLM-as-KB. This framework is universally applicable, offering value for the evaluation of a wide range of LLMs beyond those specifically analyzed in this study.

Second, our evaluation focuses primarily on factoid questions, which assess the models' ability to recall specific factual knowledge rather than perform complex reasoning. Investigating how LLMs handle such complex queries remains an important direction for future research.

\bibliography{custom}
\appendix

\section{Datasets}
\label{sec:App-data}
\texttt{SeenQA} is composed of questions selected from the following open-sourced datasets:
\begin{enumerate}
    \item Natural Questions \cite{kwiatkowski2019natural}: This dataset includes questions sourced from web queries, each paired with a corresponding Wikipedia article containing the answer. The paper on Natural Questions was submitted to TACL in April 2018.
    \item TriviaQA \cite{joshi2017triviaqa}: This dataset comprises questions from Quiz League websites, supplemented by web pages and Wikipedia searches that may contain the answer. The paper on TriviaQA was submitted to Arxiv in May 2017. For this project, we focus only on  questions supported by Wikipedia.
    \item PopQA \cite{mallen2023not}: This dataset targets long-tail entities. The authors used the Wikipedia dump from December 2018 in the retrieval augmented baseline, indicating that the knowledge in PopQA can be covered by the Wikipedia dump from that date.
\end{enumerate}
Wikipedia is a common source in the pre-training data of large language models (LLMs). Comparing the knowledge cutoff dates provided in Table~\ref{tab:llms-detail}, we can deduce that the knowledge involved in these three datasets must have been  seen during training by the LLMs used in our study.

\begin{table*}[h]
\centering
\resizebox{\textwidth}{!}{%
\begin{tabular}{cl}
\toprule
\textbf{Uninformative Type} & \multicolumn{1}{c}{\textbf{Responses Examples}} \\ \toprule
\textbf{repetition} & \begin{tabular}[c]{@{}l@{}}1. (a) (b) (c) (d) (e) (f) (g) (h) (i) (j) (k) (l) (m) (n) (o) (p) (q) (r) (s) (t) (u) (v) (w) (x) (y) (z)\\  (aa) (ab) (ac) (ad) (ae) (af) (\\ \\ 1. 2. 3. 4. 5. 6. 7. 8. 9. 10. 11. 12. 13. 14. 15. 16. 17. 18. 19. 20. 21. 22. 23. 24. 25. 26. 27.\\ \\ a swollen swollen swollen swollen swollen swollen swollen swollen swollen swollen \\ swollen swollen swollen swollen swollen swollen swollen swollen\end{tabular} \\ \hline
\textbf{none} & \begin{tabular}[c]{@{}l@{}}noah catherine cooper's mom is \_\_\_\_\_\_\_\_.\\ \\ `' \\ \\ \textless{}a href="https://i.stack.imgur.com/88888.png" rel="nofollow noreferrer"\textgreater{}\textless{}image\textgreater{}\textless{}/a\textgreater{}\end{tabular} \\ \hline
\textbf{unsure} & \begin{tabular}[c]{@{}l@{}}unsure\\ \\ i do not know.\\ \\i'm just an ai, i don't have access to real-time information or the ability to predict the future\end{tabular} \\ \bottomrule
\end{tabular}%
}
\caption{Uninformative Responses Examples}
\label{tab:unin-Re-EX}
\end{table*}

\section{LLMs Used}
\label{sec:App-llm}
Table~\ref{tab:llms-detail} summarizes the LLMs used in our experiments.
\begin{table*}[h]
\centering
\tiny
\resizebox{\textwidth}{!}{%
\begin{tabular}{c r|c c c c c c r r }
\toprule
\multirow{2}{*}{Models} & \multicolumn{1}{c}{\multirow{2}{*}{\#Params}} &  \multicolumn{1}{c}{\multirow{2}{*}{Type}} &\multicolumn{1}{c}{\multirow{2}{*}{\begin{tabular}[c]{@{}c@{}}Open\\ Source\end{tabular}}} & \multicolumn{2}{c}{Fine-Tuning} &\multicolumn{1}{c}{\multirow{2}{*}{\begin{tabular}[c]{@{}c@{}}Release\\ Date\end{tabular}}} & \multicolumn{3}{c}{Pre-Training}  \\ 
\cmidrule(lr){5-6}
\cmidrule(lr){8-10}
 & \multicolumn{1}{c}{} & \multicolumn{1}{c}{} &\multicolumn{1}{c}{} & \multicolumn{1}{c}{IT} & \multicolumn{1}{c}{RLHF} &  \multicolumn{1}{c}{} & \multicolumn{1}{c}{Knowledge} & \multicolumn{1}{c}{\# Token}   &\multicolumn{1}{c}{Vocab}   \\ \toprule
\href{https://platform.openai.com/docs/models/gpt-3-5-turbo}{gpt-3.5-turbo-0125} & Unknown & Dec-only & \usym{2717} & \usym{2713}  & \usym{2713}  & 25 Jan 2024 & Sep 2021 & - & -\\ \midrule
\multirow{5}{*}{\href{https://arxiv.org/abs/2210.11416}{Flan-T5}} & 0.08B & Enc-Dec &\usym{2713} & \usym{2713}  & \usym{2717} & 20 Oct 2022 &  \uline{April 2019} & Unknown  & 32K \\ 
 & 0.25B &Enc-Dec & \usym{2713}  & \usym{2713}  &  \usym{2717} & 20 Oct 2022 & \uline{April 2019} & Unknown & 32K \\ 
 & 0.78B &Enc-Dec &\usym{2713}  &\usym{2713}  & \usym{2717} & 20 Oct 2022 & \uline{April 2019} & Unknown & 32K \\ 
 & 3B & Enc-Dec&\usym{2713}  & \usym{2713}  &  \usym{2717} & 20 Oct 2022 & \uline{April 2019} &Unknown & 32K \\ 
 & 11B &Enc-Dec &\usym{2713}  & \usym{2713}  & \usym{2717} & 20 Oct 2022 & \uline{April 2019} & Unknown & 32K \\ \midrule
\multirow{4}{*}{\href{https://arxiv.org/abs/2302.13971}{Llama1}} & 7B & Dec-only &\usym{2713}  & \usym{2717}  & \usym{2717} & 27 Feb 2023 & \uline{Aug 2022} & 1T & 32K \\ 
 & 13B &Dec-only &\usym{2713}  &  \usym{2717} & \usym{2717} & 27 Feb 2023 & \uline{Aug 2022} & 1T & 32K \\ 
 & 65B &Dec-only &\usym{2713}  &  \usym{2717} & \usym{2717} & 27 Feb 2023 & \uline{Aug 2022} & 1.4T & 32K \\ \midrule
\multirow{3}{*}{\href{https://arxiv.org/abs/2307.09288}{Llama2}} & 7B &Dec-only &\usym{2713}  & \usym{2717} & \usym{2717} & 18 July 2023 & Sep 2022 & 2T & 32K  \\ 
 & 13B & Dec-only&\usym{2713}  & \usym{2717} &\usym{2717}  & 18 July 2023 & Sep 2022 & 2T & 32K   \\ \vspace{1.5mm} 
 & 70B & Dec-only&\usym{2713}  & \usym{2717} & \usym{2717} & 18 July 2023 & Sep 2022 & 2T & 32K \\ 
\multirow{3}{*}{\href{https://arxiv.org/abs/2307.09288}{Llama2chat}} & 7B & Dec-only&\usym{2713}  & \usym{2713}  & \usym{2713}  & 18 July 2023 & Sep 2022 & 2T & 32K  \\ 
 & 13B &Dec-only &\usym{2713}  & \usym{2713}  & \usym{2713}   & 18 July 2023 & Sep 2022 & 2T & 32K  \\ 
 & 70B & Dec-only&\usym{2713}  & \usym{2713}  & \usym{2713}  & 18 July 2023& Sep 2022 & 2T & 32K  \\ \midrule
\multirow{2}{*}{\href{https://ai.meta.com/blog/meta-llama-3/}{Llama3}} & 8B & Dec-only&\usym{2713}  & \usym{2717}  & \usym{2717} & 18 April 2024& Mar 2023 & 15T+ & 128K    \\ \vspace{1.5mm} 
 & 70B &Dec-only &\usym{2713}  & \usym{2717}  & \usym{2717} & 18 April 2024  & Dec 2023 & 15T+ & 128K   \\ 
\multirow{2}{*}{\href{https://ai.meta.com/blog/meta-llama-3/}{Llama3Instruct}} & 8B & Dec-only&\usym{2713}   & \usym{2713}  & \usym{2713} & 18 April 2024  & Mar 2023  & 15T+ & 128K  \\ 
 & 70B &Dec-only &\usym{2713}  & \usym{2713}  & \usym{2713}  & 18 April 2024  & Dec 2023 & 15T+ & 128K   \\ \midrule \vspace{1.5mm} 
\href{https://arxiv.org/abs/2310.06825}{Mistral} & 7B &  Dec-only&\usym{2713} & \usym{2717} &\usym{2717}  & 27 Sep 2023 & Unknown & Unknown & 32K \\ 
\href{https://arxiv.org/abs/2310.06825}{Mistral-Instruct} & 7B & Dec-only&\usym{2713}  & \usym{2713} & \usym{2717} & 27 Sep 2023  & Unknown & Unknown & 32K \\ \midrule
\multirow{2}{*}{\href{https://arxiv.org/abs/2403.08295}{Gemma}} & 2B & Dec-only&\usym{2713} & \usym{2717}  & \usym{2717}  & 21 Feb 2024  & Unknown & 3T & 256K\\ \vspace{1.5mm} 
 & 7B & Dec-only&\usym{2713} & \usym{2717}  & \usym{2717}  & 21 Feb 2024 & Unknown & 6T & 256K \\ 
\multirow{2}{*}{\href{https://arxiv.org/abs/2403.08295}{Gemma-Instruct}} & 2B & Dec-only &\usym{2713} & \usym{2713}  & \usym{2713} & 21 Feb 2024  & Unknown & 3T & 256K  \\ 
 & 7B & Dec-only &\usym{2713} & \usym{2713}  & \usym{2713} & 21 Feb 2024& Unknown & 6T & 256K\\ \midrule
\href{https://www.microsoft.com/en-us/research/blog/phi-2-the-surprising-power-of-small-language-models/}{Phi2} & 3B & Dec-only&\usym{2713} & \usym{2717}  & \usym{2717} & 12 Dec 2023 & Unknown & 1.4T & 50K  \\ \bottomrule
\end{tabular}%
}
\caption{Summary of LLMs used in our experiments. `IT' denotes Instruction Tuning, and `RLHF' refers to Reinforcement Learning from Human Feedback. `Knowledge' indicates the knowledge cutoff date. Underlined dates were not explicitly provided by the authors but extrapolated from the datasets used for LLM training. Flan-T5's base model is T5 version 1.1 pre-trained on the C4 dataset, filtered from web-extracted text in April 2019. Llama 1's pre-training data includes Wikipedia dumps from June to August 2022.}
\label{tab:llms-detail}
\end{table*}

\section{Uninformative Responses Examples}
\label{sec:App-unin-EX}
\begin{figure}[h]
    \centering
    \includegraphics[width=.48\textwidth]{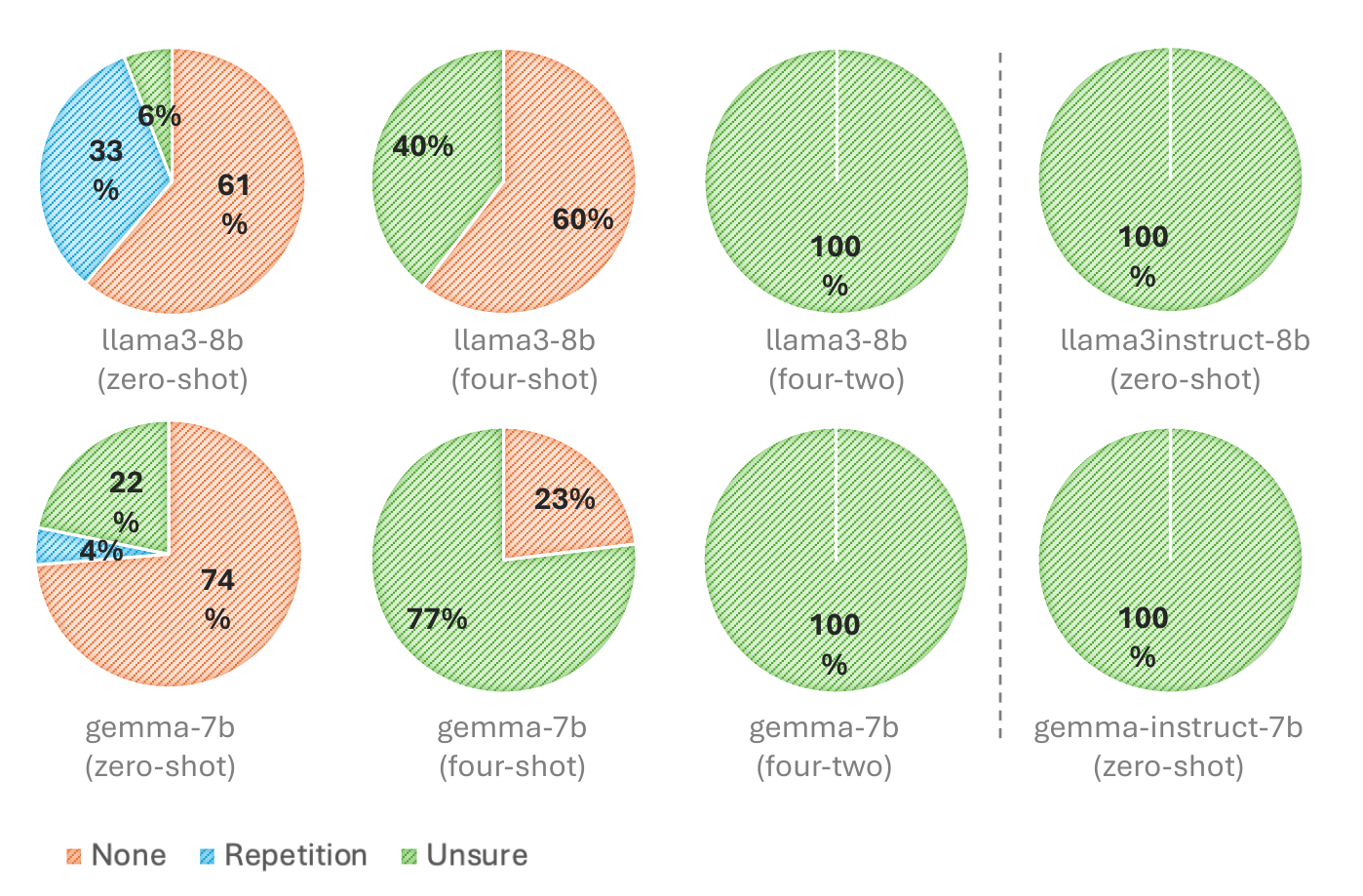}
    \caption{Distribution of uninformative responses given by LLMs to questions about unseen knowledge. We report results for the  \textsc{llama3-8b}, \textsc{gemma-7b}, and their fine-tuned models (fourth column) but observe similar trends on other models (omitted for the sake of brevity).}
    \label{fig:nr-dist}
\end{figure}

Table~\ref{tab:unin-Re-EX} provides some examples of uninformative responses. As illustrated in Figure~\ref{fig:nr-dist} (fourth column), fine-tuned LLMs are able to explicitly acknowledge their lack of knowledge  by answering `unsure' to questions about unseen knowledge. In contrast, base LLMs often produce responses classified as `none' or `repetition' in the absence of unsure shots (contrast columns one and two with column three in Figure~\ref{fig:nr-dist}).

\section{Prompts Used}
\label{sec:App-prompt}
To provide a comprehensive evaluation, we experimented with three types of prompt settings: zero-shot, four-shot, and four-shot with two unsure shots.  
To avoid any bias introduced by fixed examples, we employed a dynamic few-shot method following the work of \citet{nori2023can}.
We collected two repositories, $R_{seen}$ and $R_{unseen}$. 
$R_{seen}$ includes 280 question-answer pairs about seen knowledge (200 from the unused data of PopQA and training data of Natural Questions and TriviaQA; 80 are generated using the templates in Table~\ref{tab: temps}). 
$R_{unseen}$ consists of 40 question-answer pairs about unseen knowledge, all generated using the templates in Table~\ref{tab: temps}. 
We used \href{https://platform.openai.com/docs/models/embeddings}{\textsc{text-embedding-3-small}}
 to embed the questions in the repositories and test questions as vector representations.  
 For each test question under the four-shot setting, we retrieved its nearest four questions from  $R_{seen}$.
 Under the four-shot with two unsure shots setting, we retrieved the nearest two questions from $R_{seen}$ and two from $R_{unseen}$.

The prompt used to detect time-sensitive questions is shown in Table~\ref{tab:prompt_time_sen}.
The QA prompts under three different prompt settings are shown in Table~\ref{tab:prompt_qa}.
The prompt used to check whether an LLM's response matches the ground-truth is shown in Table~\ref{tab:correct_check}.
The prompt used to generate distractors for the consistency test is shown in Table~\ref{tab:distractor}.
\begin{table*}[h]
\centering
\small
\begin{tabular}{p{.95\textwidth}}       
\hline
\textbf{Prompt for detecting time-sensitive questions} \\
\hline
INSTRUCTION: Please provide the index of questions whose answers change yearly. Just return the index without explanations. \\
\\

Here is the list of questions:\\
1. Who is the most paid player in EPL?\\
2. What is the capital of Louisiana?\\
3. Who won the Nobel Peace Prize in 2009?\\
4. What is the latest model of the iPhone currently available?\\
Index:\\
1, 4\\
\\

Here is the list of questions:\\
{[}question placeholder{]}\\
Index:\\

\hline
\end{tabular}
\caption{The prompt for detecting time-sensitive questions }
\label{tab:prompt_time_sen}
\end{table*}

\begin{table*}[h]
\centering
\small
\begin{tabular}{p{.95\textwidth}}       
\hline
\textbf{QA prompt in zero-shot} \\
\hline
INSTRUCTION: Please answer knowledge-related questions directly. Note: Please do not give anything other than the answer; Say "unsure" if you do not know.\\
\\

QUESTION: {[}question placeholder{]}\\
ANSWER:\\
\hline
\textbf{QA prompt in four-shot} \\
\hline
INSTRUCTION: Please answer knowledge-related questions directly. Note: Please do not give anything other than the answer; Say "unsure" if you do not know.\\
\\

QUESTION: {[}question example 1 from $R_{seen}${]}\\
ANSWER: {[}answer 1{]}\\
\\

QUESTION: {[}question example 2 from $R_{seen}${]}\\
ANSWER: {[}answer 2{]}\\
\\

QUESTION: {[}question example 3 from $R_{seen}${]}\\
ANSWER: {[}answer 3{]}\\
\\

QUESTION:  {[}question example 4 from $R_{seen}${]}\\
ANSWER: {[}answer 4{]}\\
\\

QUESTION: {[}question placeholder{]}\\
ANSWER:\\
\hline
\textbf{QA prompt in four-shot with tew unsure shot} \\
\hline
INSTRUCTION: Please answer knowledge-related questions directly. Note: Please do not give anything other than the answer; Say "unsure" if you do not know.\\
\\

QUESTION: {[}question example 1 from $R_{seen}${]}\\
ANSWER: {[}answer 1{]}\\
\\

QUESTION: {[}question example 2 from $R_{seen}${]}\\
ANSWER: {[}answer 2{]}\\
\\

QUESTION: {[}question example 3 from $R_{unseen}${]}\\
ANSWER: unsure\\
\\

QUESTION:  {[}question example 4 from $R_{unseen}${]}\\
ANSWER: unsure\\
\\

QUESTION: {[}question placeholder{]}\\
ANSWER:\\
\hline
\end{tabular}
\caption{The question answering prompt format. The shots are selected from repositories, $R_{seen}$ and $R_{seen}$. The order of shots is random. For the MCQ tests in consistency experiments, we edit the instruction line to \textit{INSTRUCTION: Please answer knowledge-related multi-choice questions directly. Note: Please do not give anything other than the appropriate option (A, B, C, D or E); choose the option indicating "unsure" if you do not know.}}
\label{tab:prompt_qa}
\end{table*}


\begin{table*}[h]
\centering
\small
\begin{tabular}{p{.95\textwidth}}       
\hline
\textbf{Prompt for check whether an answer matches the ground truth for the question} \\
\hline
INSTRUCTION: You need to check whether the prediction of a question-answering system to a question is correct. You should make the judgment based on a list of ground truth answers provided to you. Your response should be "yes" if the prediction is correct or "no" if the prediction is wrong.\\
\\

Question: Who authored The Taming of the Shrew (published in 2002)? \\
Ground truth: ["William Shakespeare", "Roma Gill"] \\
prediction: W Shakespeare \\
Correctness: yes\\

Question: What country is Maharashtra Metro Rail Corporation Limited located in?\\
Ground truth: ["India"] \\
prediction: Maharashtra \\
Correctness: no\\
\\

Question: Edward Tise (known for Full Metal Jacket (1987)) is in what department? \\
Ground truth: ["sound department"] \\
Prediction: 2nd Infantry Division, United States Army \\
Correctness: no\\
\\

Question: Which era did Michael Oakeshott belong to? \\
Ground truth: ["20th-century philosophy"] \\
prediction: 20th century. \\
Correctness: yes \\
\\

Question: {[}quesetion placeholder{]}\\
Ground truth: {[}ground truth placeholder{]}\\
prediction: {[}LLM's answer placeholder{]}\\
Correctness:\\
\hline
\end{tabular}
\caption{The prompt used to check whether an LLM's answer matches the ground truth for the question}
\label{tab:correct_check}
\end{table*}


\begin{table*}[h]
\centering
\small
\begin{tabular}{p{.95\textwidth}}       
\hline
\textbf{Prompt for generating distractors} \\
\hline
INSTRUCTION: For the given question-answer pair, provide 20 different distractors that are similar yet distinct from the given answer. Note: Seperate the 20 distractors with a special token "[SEP]". \\
\\

Q: Who was the President of the United States in 2010?\\
A: Barack Obama\\
Distractors: George W. Bush [SEP] Bill Clinton [SEP] Ronald Reagan [SEP] Donald Trump [SEP] Jimmy Carter [SEP] George H.W. Bush [SEP] Richard Nixon [SEP] Gerald Ford [SEP] Lyndon B. Johnson [SEP] John F. Kennedy [SEP] Dwight D. Eisenhower [SEP] Harry S. Truman [SEP] Franklin D. Roosevelt [SEP] Herbert Hoover [SEP] Calvin Coolidge [SEP] Woodrow Wilson [SEP] William Howard Taft [SEP] Theodore Roosevelt [SEP] William McKinley [SEP] Grover Cleveland\\
\\

Q: What is the name of the first cloned sheep?\\
A: the first cloned sheep is dolly.\\
Distractors: the first cloned sheep is Polly [SEP] the first cloned sheep is Molly [SEP] the first cloned sheep is Holly [SEP] the first cloned sheep is Bella [SEP] the first cloned sheep is Daisy [SEP] the first cloned sheep is Lily [SEP] the first cloned sheep is Rosie [SEP] the first cloned sheep is Millie [SEP] the first cloned sheep is Ellie [SEP] the first cloned sheep is Sally [SEP] the first cloned sheep is Tilly [SEP] the first cloned sheep is Nelly [SEP] the first cloned sheep is Jolly [SEP] the first cloned sheep is Betty [SEP] the first cloned sheep is Annie [SEP] the first cloned sheep is Lucy [SEP] the first cloned sheep is Maggie [SEP] the first cloned sheep is Cindy [SEP] the first cloned sheep is Penny [SEP] the first cloned sheep is Ginny\\

\\
Q: [QUESTION]\\
A: [ANSWER]\\
Distractors:\\
\hline
\end{tabular}
\caption{The prompt used to generate distractors for consistency tests.}
\label{tab:distractor}
\end{table*}


\section{Full Experimental Results}
\label{sec:App-ex}
\begin{figure*}[t]
    \centering
    \includegraphics[width=\textwidth]{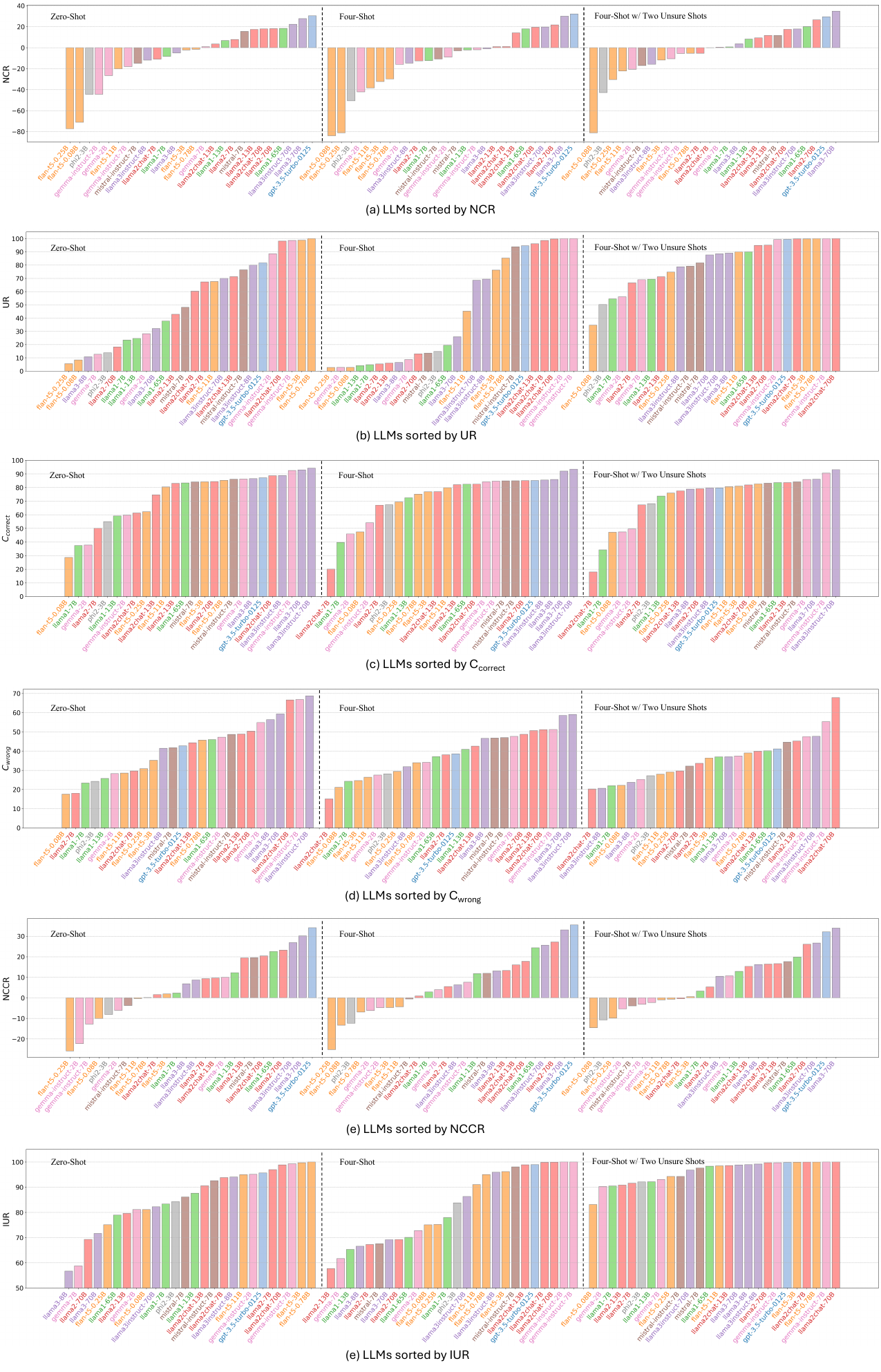}
    \caption{Ranking of LLMs based on different Metrics. }
    \label{fig:rank}
\end{figure*}

\begin{figure*}[t]
    \centering
    \includegraphics[width=\textwidth]{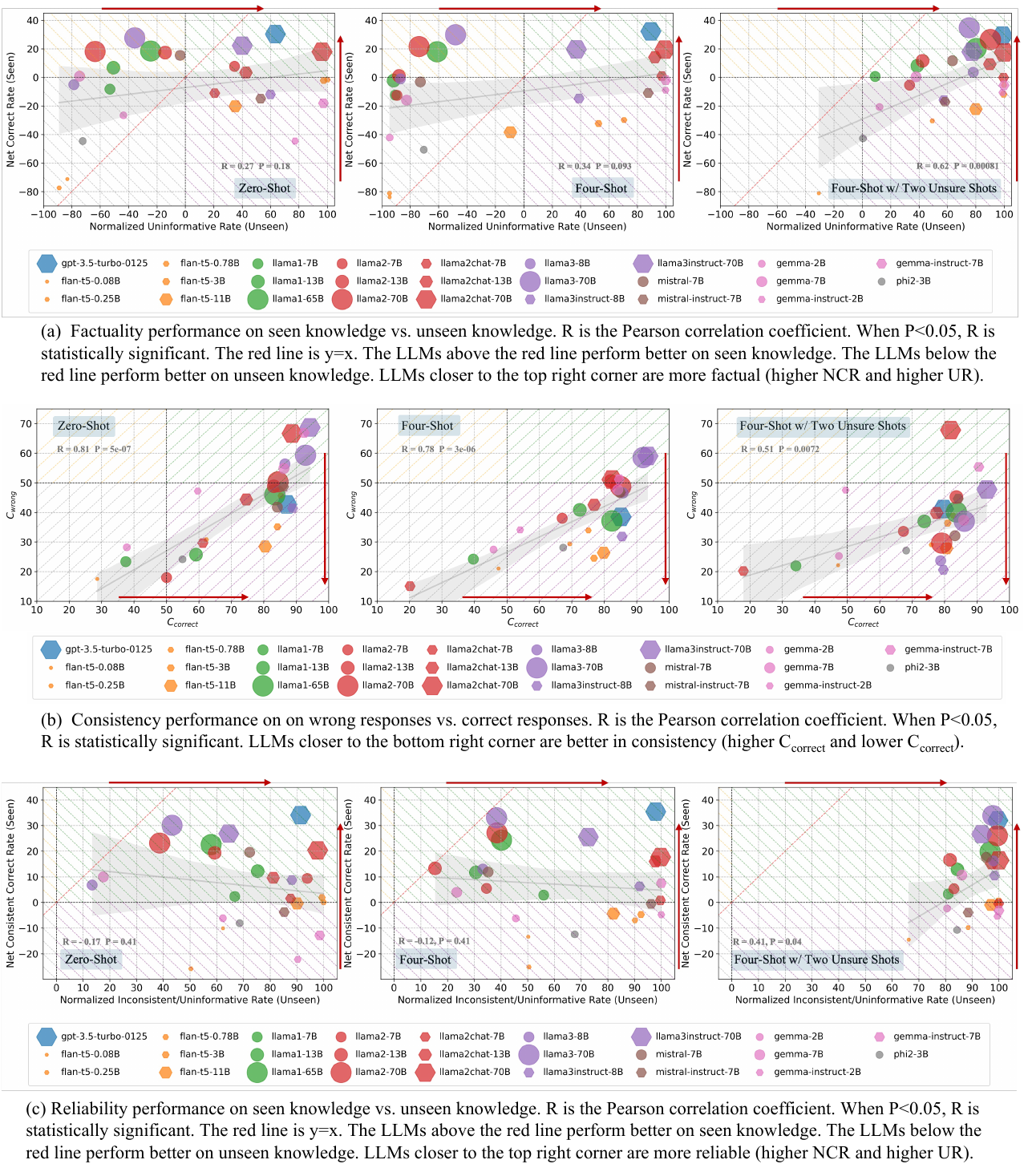}
    \caption{Visualization of LLMs' factuality, consistency and reliablity performance.}
    \label{fig:fcr}
\end{figure*}

Table~\ref{tab:factulity}, Table~\ref{tab:consistency}, and Table~\ref{tab:reliability} provides the detailed results of LLMs' performance on factuality, consitency, and reliablity  respectively.
Figure~\ref{fig:rank} shows the rankings of LLMs based on different metrics.
Figure~\ref{fig:fcr} compare different LLMs' factuality, consistency, and reliability performance.

\begin{figure*}[h]
    \centering
    \includegraphics[width=\textwidth]{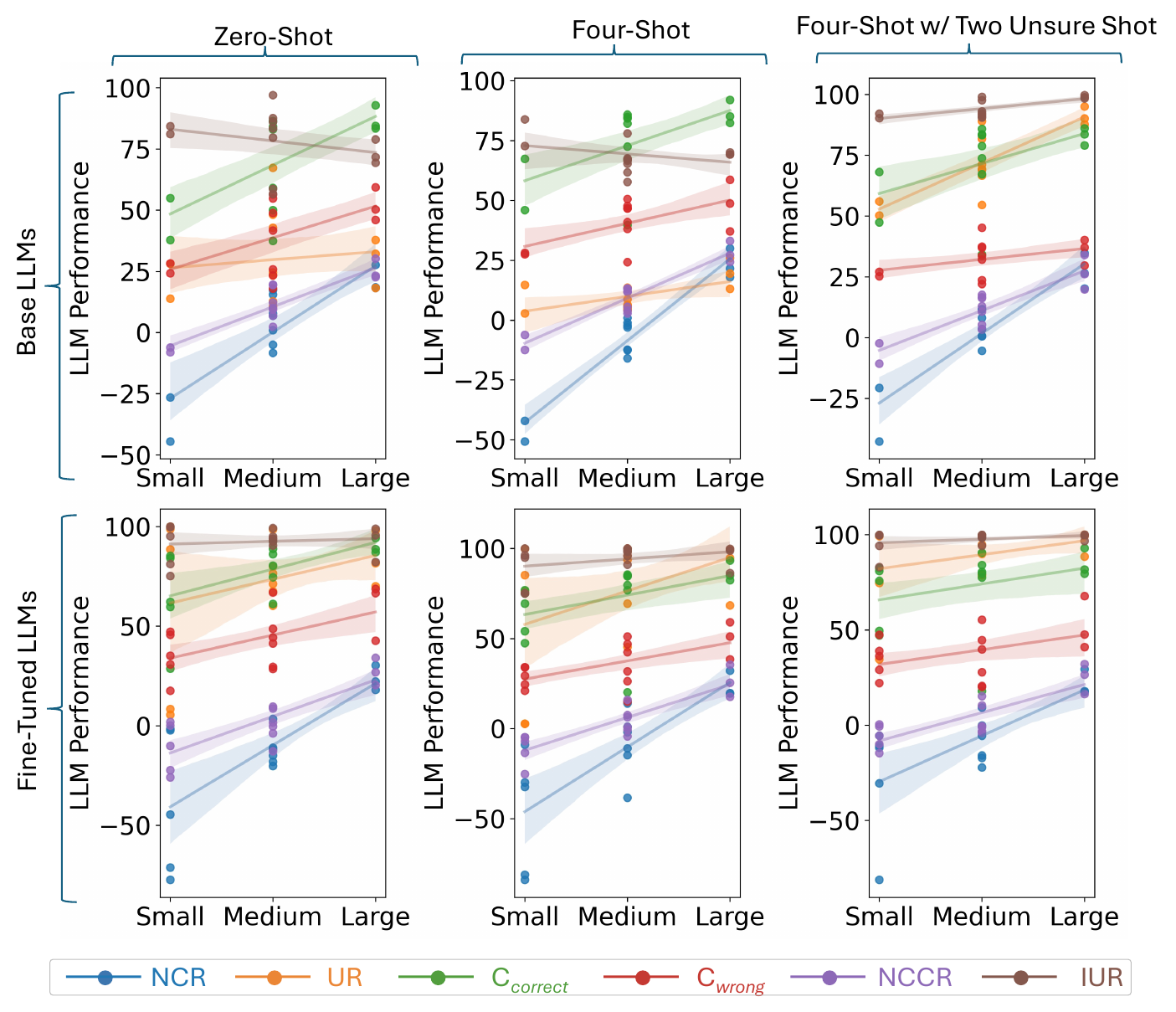}
    \caption{The impact of model size on LLM performance, measured with NCR, UR, C$_{correct}$, C$_{wrong}$, NCCR, and IUR (values are scaled by 100).   Different metrics are color-coded. LLMs are shown in three sizes, small, medium, and large and are grouped into `base' and fine-tuned ones. }
    \label{fig:modelSize}
\end{figure*}

\begin{figure*}[h]
    \centering
    \includegraphics[width=\textwidth]{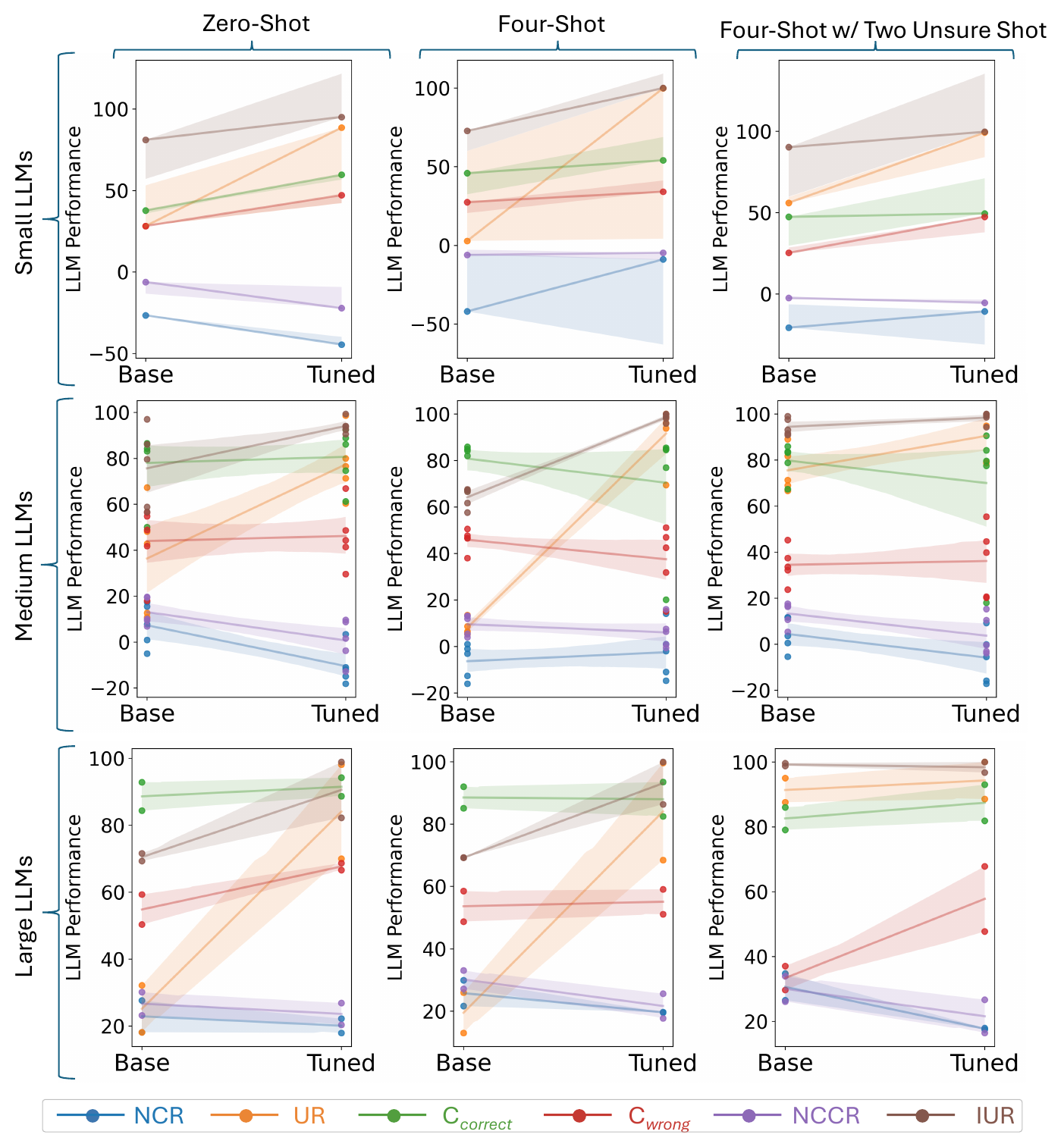}
    \caption{The impact of fine-tuning on LLM performance, measured with NCR, UR, C$_{correct}$, C$_{wrong}$, NCCR, and IUR (values are scaled by 100).  Different metrics color-coded. This analysis only considers the performance of  Llama2, Llama3, Mistral, and Gemma as these families include both base LLMs and  fine-tuned versions. Models are shown in three sizes, small, medium, and large.}
    \label{fig:fineTuned}
\end{figure*}

\begin{figure*}[h]
    \centering
    \includegraphics[width=\textwidth]{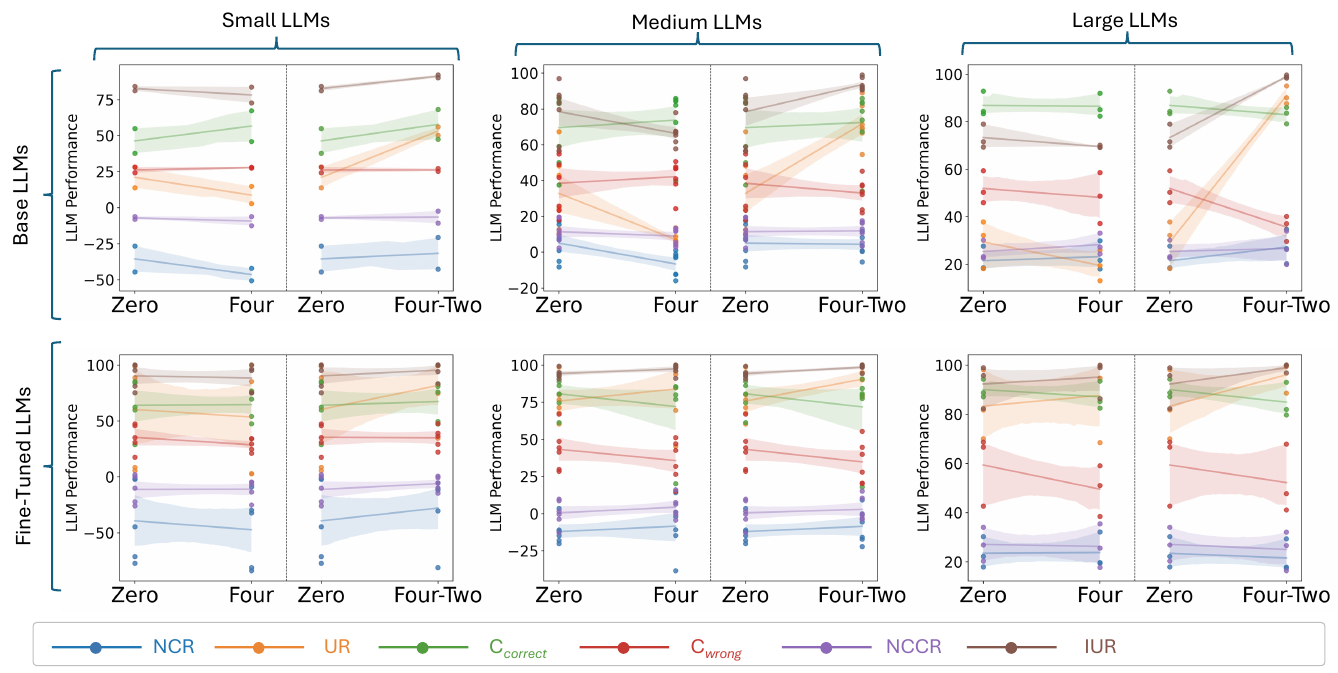}
    \caption{The impact of ICL on LLM performance, measured with NCR, UR, C$_{correct}$, C$_{wrong}$, NCCR, and IUR (values are scaled by 100). Different metrics are color-coded. We compare zero-shot and four-shot settings;  and zero-shot against four-shot with two unsure shots. LLMs (`base' and fine-tuned ones) are in three sizes, small, medium, and large.}
    \label{fig:ICL}
\end{figure*}

Figure~\ref{fig:modelSize}, ~\ref{fig:fineTuned}, and ~\ref{fig:ICL} provide detailed illustration of the impact of model size, fine-tuning, and ICL.

\begin{figure*}[h]
    \centering
    \includegraphics[width=\textwidth]{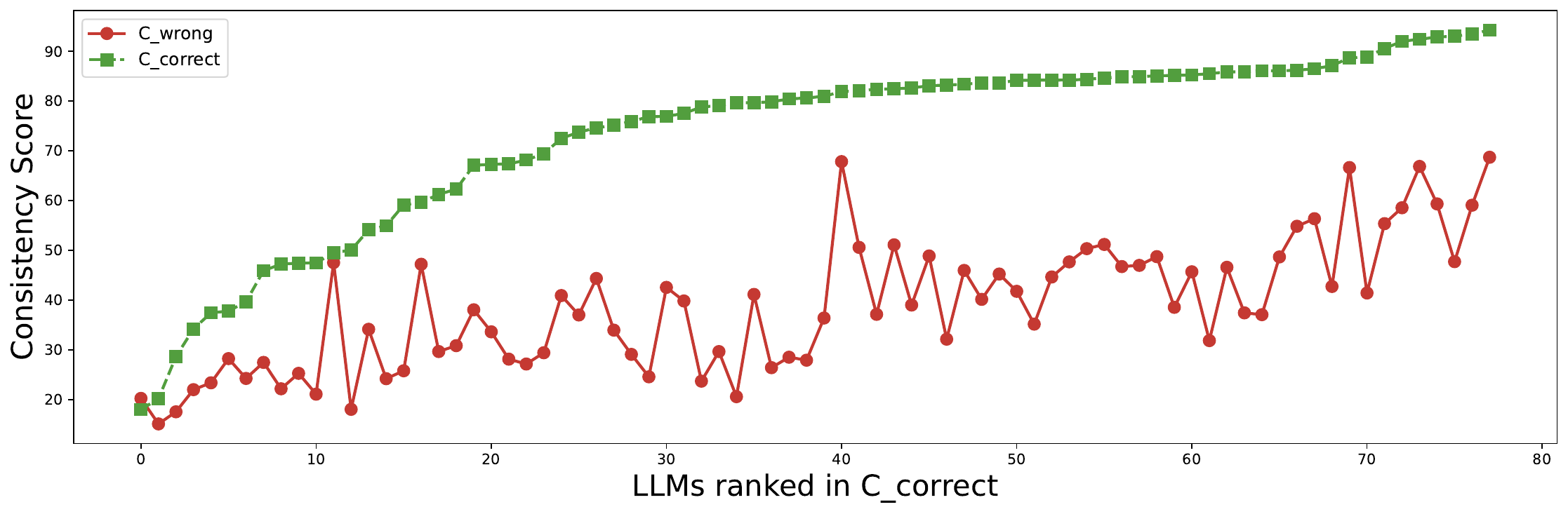}
    \caption{$C_{\textit{correct}}$ and $C_{\textit{wrongs}}$ performance of LLMs ranked in $C_{\textit{correct}}$. }
    \label{fig:C_correct_and_C_wrong}
\end{figure*}

Figure~\ref{fig:C_correct_and_C_wrong} shows $C_{\textit{correct}}$ and $C_{\textit{wrongs}}$ performance of LLMs ranked in $C_{\textit{correct}}$.

\begin{figure*}[h]
    \centering
    \includegraphics[width=\textwidth]{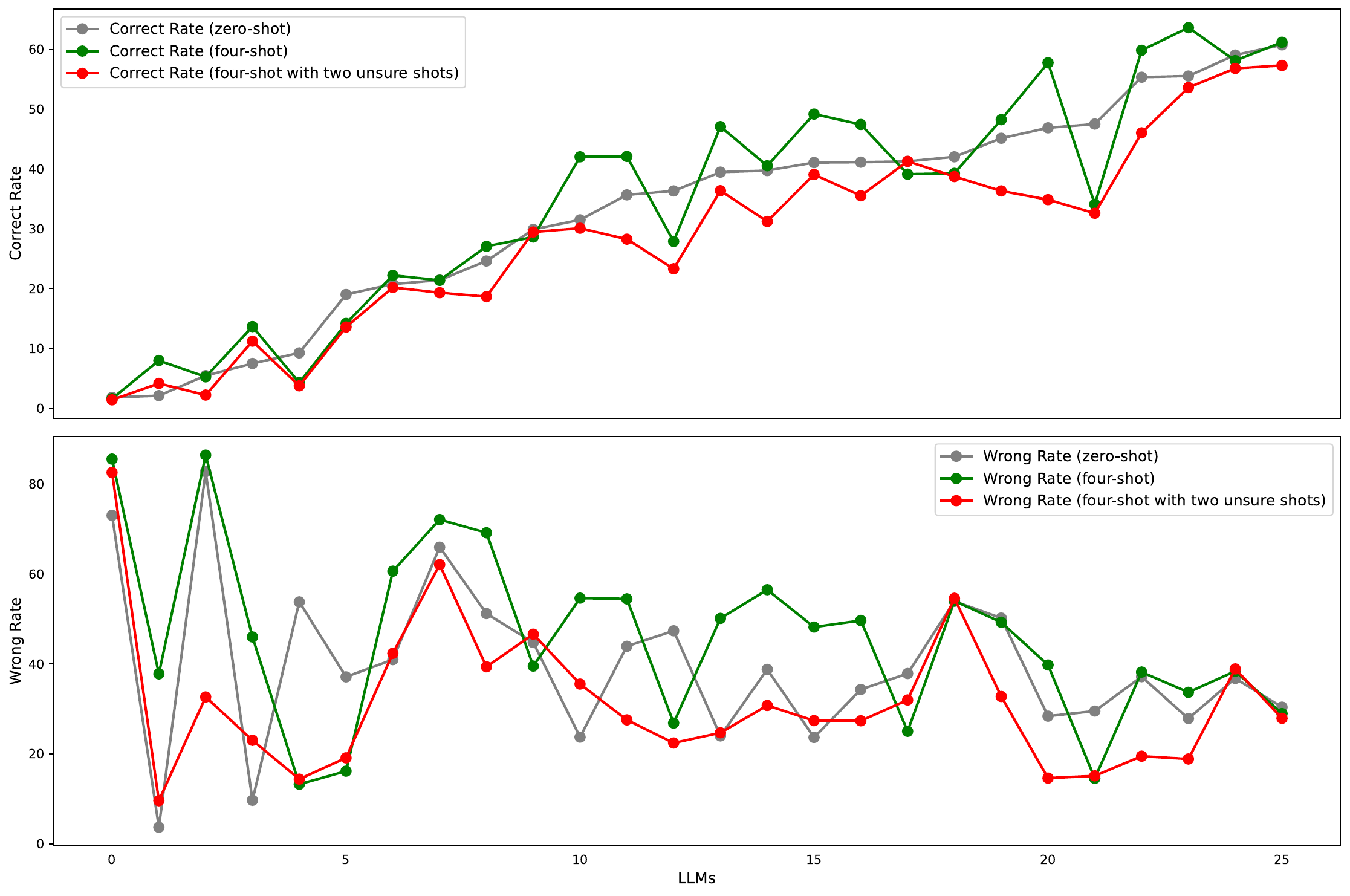}
    \caption{Correct Rate and Wrong Rate under different prompt settings for LLMs ranked in Correct Rate under zero-shot.}
    \label{fig:CRNR}
\end{figure*}

Figure~\ref{fig:CRNR} shows correct rate and wrong ate under different prompt settings for LLMs ranked in correct rate under zero-shot.

\begin{table*}[h]
\centering

\resizebox{.7\textwidth}{!}{%
\begin{tabular}{cr|rr>{\columncolor[HTML]{E9EFF6}}r>{\columncolor[HTML]{E9EFF6}}r|rr>{\columncolor[HTML]{E9EFF6}}r>{\columncolor[HTML]{E9EFF6}}r|rr>{\columncolor[HTML]{E9EFF6}}r>{\columncolor[HTML]{E9EFF6}}r}
\toprule
\multirow{2}{*}{\textbf{Model}} & \multicolumn{1}{c}{\multirow{2}{*}{\textbf{Params}}} & \multicolumn{4}{c}{\textbf{zero-shot}} & \multicolumn{4}{c}{\textbf{four-shot}} & \multicolumn{4}{c}{\textbf{four-shot-2}} \\
\cmidrule(lr){3-6}
\cmidrule(lr){7-10}
\cmidrule(lr){11-14}
 & \multicolumn{1}{c}{} 
 & \multicolumn{3}{c}{\textbf{Seen}} 
 & \multicolumn{1}{c}{\textbf{Uneen}}  
 & \multicolumn{3}{c}{\textbf{Seen}} 
 & \multicolumn{1}{c}{\textbf{Uneen}}  
 & \multicolumn{3}{c}{\textbf{Seen}} 
 & \multicolumn{1}{c}{\textbf{Uneen}} \\
 \cmidrule(lr){3-5}
  \cmidrule(lr){6-6}
\cmidrule(lr){7-9}
\cmidrule(lr){10-10}
\cmidrule(lr){11-13}
\cmidrule(lr){14-14}
 & \multicolumn{1}{c}{} 
 & \multicolumn{1}{c}{\textbf{WR ($\downarrow$)}} 
 & \multicolumn{1}{c}{\textbf{CR ($\uparrow$)}} 
 & \multicolumn{1}{c}{\cellcolor[HTML]{E9EFF6}\textbf{NCR ($\uparrow$)}}
 & \multicolumn{1}{c}{\cellcolor[HTML]{E9EFF6}\textbf{UR ($\uparrow$)}} 
 & \multicolumn{1}{c}{\textbf{WR ($\downarrow$)}} 
 & \multicolumn{1}{c}{\textbf{CR ($\uparrow$)}} 
 & \multicolumn{1}{c}{\cellcolor[HTML]{E9EFF6}\textbf{NCR ($\uparrow$)}} 
 & \multicolumn{1}{c}{\cellcolor[HTML]{E9EFF6}\textbf{UR ($\uparrow$)}} 
 & \multicolumn{1}{c}{\textbf{WR ($\downarrow$)}} 
 & \multicolumn{1}{c}{\textbf{CR ($\uparrow$)}} 
 & \multicolumn{1}{c}{\cellcolor[HTML]{E9EFF6}\textbf{NCR ($\uparrow$)}}
 & \multicolumn{1}{c}{\cellcolor[HTML]{E9EFF6}\textbf{UR ($\uparrow$)}} \\
 \toprule
  GPT-3.5 Turbo & Unknown  & 30.40 & 60.73 & 30.33 & 81.70 & 28.97 & 61.17 & 32.20 & 94.70 &  27.93 & 57.30 & 29.37 & 99.37 \\ \midrule
\multirow{5}{*}{Flan-T5} & 0.08B & 73.03 & 1.83 & -71.20 & 8.40 & 85.53 & 1.63 & -83.90 & 2.77 & 82.57 & 1.43 & -81.13 & 34.63 \\
 & 0.25B & 82.77 & 5.47 & -77.30 & 5.50 & 86.43 & 5.27 & -81.17 & 2.70 & 32.67 & 2.23 & -30.43 & 74.67\\
 & 0.78B & 3.70 & 2.13 & -1.57 & 100.00 & 37.80 & 8.00 & -29.80 & 85.30 & 9.60 & 4.17 & -5.43 & 99.90\\
 & 3B & 9.70 & 7.50 & -2.20 & 98.73 & 46.00 & 13.67 & -32.33 & 76.27 & 23.03 & 11.23 & -11.80 & 99.73\\
 & 11B  & 40.97 & 20.77 & -20.20 & 67.57 & 60.63 & 22.23 & -38.40 & 45.27 & 42.37 & 20.20 & -22.17 & 90.00\\ \midrule
\multirow{4}{*}{Llama 1} & 7B  & 43.93 & 35.67 & -8.27 &23.40 & 54.47 & 42.10 & -12.37 & 4.73& 27.57 & 28.27 & 0.70 &  54.50\\
 & 13B  & 34.33 & 41.13 & 6.80 &24.63 & 49.67 & 47.43 & -2.23 & 4.10& 27.37 & 35.53 & 8.17 & 69.37\\
 & 65B & 28.40 & 46.87 & 18.47 &37.77 & 39.77 & 57.73 & 17.97 & 19.47 & 14.63 & 34.87 & 20.23 & 90.10\\  \midrule
\multirow{3}{*}{Llama 2} & 7B  & 23.73 & 31.50 & 7.77 & 67.30 & 54.63 & 42.03 & -12.60 & 5.33 & 35.53 & 30.10 & -5.43 & 66.60 \\
 & 13B  & 23.67 & 41.07 & 17.40 & 42.83 & 48.20 & 49.17 & 0.97 & 6.03 & 27.40 & 39.07 & 11.67 & 71.23\\ \vspace{1.5mm} 
 & 70B  & 37.17 & 55.33 & 18.17 & 18.17 & 38.20 & 59.83 & 21.63 & 13.03 & 19.50 & 46.03 & 26.53 & 95.10\\ 
\multirow{3}{*}{Llama2chat} & 7B & 47.37 & 36.33 & -11.03 & 60.30 & 26.87 & 27.90 & 1.03 & 98.60 & 23.43 & 23.33 & -0.10 &  99.73\\
 & 13B & 37.87 & 41.27 & 3.40 & 71.30 & 25.03 & 39.13 & 14.10 & 96.13 & 32.00 & 41.27 & 9.27 & 94.90\\ 
 & 70B  & 29.53 & 47.50 & 17.97 & 98.10 & 14.57 & 34.10 & 19.53 & 99.63 & 15.13 & 32.60 & 17.47 & 100.00\\  \midrule
\multirow{2}{*}{Llama3} & 8B  & 50.20 & 45.13 & -5.07 & 10.73 & 49.27 & 48.23 & -1.03 & 6.50 & 32.77 & 36.33 & 3.57 & 89.03\\ \vspace{1.5mm} 
 & 70B & 27.87 & 55.53 & 27.67 & 32.13 & 33.70 & 63.60 & 29.90 & 25.93 & 18.87 & 53.60 & 34.73 &  87.63\\
\multirow{2}{*}{Llama3Instruct} & 8B & 53.93 & 42.03 & -11.90 & 79.97 & 54.00 & 39.27 & -14.73 & 69.43 & 54.60 & 38.73 & -15.87 & 78.73\\
 & 70B  & 36.80 & 59.03 & 22.23 & 70.03 & 38.40 & 58.10 & 19.70 & 68.47 & 38.90 & 56.80 & 17.90 & 88.60\\ \midrule \vspace{1.5mm} 
Mistral & 7B  & 24.00 & 39.47 & 15.47 & 48.13 & 50.13 & 47.07 & -3.07 & 13.57 & 24.70 & 36.37 & 11.67 & 81.73 \\ 
Mistral-Instruct & 7B & 44.77 & 29.90 & -14.87 & 76.50 & 39.53 & 28.63 & -10.90 & 93.80 & 46.63 & 29.47 & -17.17 & 79.13\\ \midrule
\multirow{2}{*}{Gemma} & 2B  & 51.20 & 24.63 & -26.57 & 28.17 & 69.17 & 27.07 & -42.10 & 2.77 & 39.37 & 18.67 & -20.70 & 56.07\\ \vspace{1.5mm} 
 & 7B & 38.80 & 39.73 & 0.93 & 12.70 & 56.50 & 40.53 & -15.97 & 8.67 &30.77 & 31.23 & 0.47 & 68.93\\
\multirow{2}{*}{Gemma-Instruct} & 2B  & 53.80 & 9.27 & -44.53 & 88.60 & 13.27 & 4.30 & -8.97 & 99.93 & 14.40 & 3.77 & -10.63 & 99.30\\ 
 & 7B  & 37.13 & 19.03 & -18.10 & 98.60 & 16.17 & 14.20 & -1.97 & 99.97 & 19.13 & 13.60 & -5.53 & 99.93\\ \midrule 
Phi2 & 3B  & 65.97 & 21.43 & -44.53 & 13.83 & 72.10 & 21.40 & -50.70 & 14.77 & 62.07 & 19.33 & -42.73 &  50.20\\ 
\bottomrule
\end{tabular}%
}
\caption{Factuality performance (Values are scaled by 100).}
\label{tab:factulity}
\end{table*}

\begin{table*}[h]
\centering

\resizebox{.7\textwidth}{!}{%
\begin{tabular}{cr|rr>{\columncolor[HTML]{E9EFF6}}r>{\columncolor[HTML]{E9EFF6}}r|rr>{\columncolor[HTML]{E9EFF6}}r>{\columncolor[HTML]{E9EFF6}}r|rr>{\columncolor[HTML]{E9EFF6}}r>{\columncolor[HTML]{E9EFF6}}r}
\toprule
\multirow{2}{*}{\textbf{Model}} & \multicolumn{1}{c}{\multirow{2}{*}{\textbf{Params}}} & \multicolumn{4}{c}{\textbf{zero-shot}} & \multicolumn{4}{c}{\textbf{four-shot}} & \multicolumn{4}{c}{\textbf{four-shot-2}} \\
\cmidrule(lr){3-6}
\cmidrule(lr){7-10}
\cmidrule(lr){11-14}
 & \multicolumn{1}{c}{} & \multicolumn{1}{c}{\textbf{ $C_{\textit{wrong}}^{s}$($\downarrow$)}} & \multicolumn{1}{c}{\textbf{ $C_{\textit{wrong}}^{u}$($\downarrow$)}} & \multicolumn{1}{c}{\cellcolor[HTML]{E9EFF6}\textbf{ $C_{\textit{wrong}}$($\downarrow$)}} & \multicolumn{1}{c}{\cellcolor[HTML]{E9EFF6}\textbf{$C_{\textit{correct}}$ ($\uparrow$)}} & \multicolumn{1}{c}{\textbf{ $C_{\textit{wrong}}^{s}$($\downarrow$)}} & \multicolumn{1}{c}{\textbf{ $C_{\textit{wrong}}^{u}$($\downarrow$)}} & \multicolumn{1}{c}{\cellcolor[HTML]{E9EFF6}\textbf{$C_{\textit{wrong}}$ ($\downarrow$)}} & \multicolumn{1}{c}{\cellcolor[HTML]{E9EFF6}\textbf{$C_{\textit{correct}}$ ($\uparrow$)}} & \multicolumn{1}{c}{\textbf{ $C_{\textit{wrong}}^{s}$($\downarrow$)}} & \multicolumn{1}{c}{\textbf{ $C_{\textit{wrong}}^{u}$($\downarrow$)}} & \multicolumn{1}{c}{\textbf{\cellcolor[HTML]{E9EFF6}$C_{\textit{wrong}}$ ($\downarrow$)}} & \multicolumn{1}{c}{\cellcolor[HTML]{E9EFF6}\textbf{$C_{\textit{correct}}$ ($\uparrow$)}}  \\
 \toprule
  GPT-3.5 Turbo & -  & 61.79 & 23.65 & 42.72 & 87.10 & 57.43 & 19.62 & 38.53 & 85.16 & 48.56 & 33.68 & 41.12 & 79.68 \\ \midrule
\multirow{5}{*}{Flan-T5} & 0.08B & 14.49 & 20.56 & 17.53 & 28.64 & 16.53 & 25.62 & 21.07 & 47.45 & 18.44 & 25.87 & 22.15 & 47.21\\
 & 0.25B  & 35.33 & 26.31 & 30.82 & 62.29 & 33.36 & 25.44 & 29.40 & 69.40 & 35.34 & 22.80 & 29.07 & 75.90 \\
 & 0.78B  & 45.68 & - & 45.68 & 85.23 & 34.16 & 33.72 & 33.94 & 75.12 & 42.99 & 35.00 & 38.99 & 82.64 \\
 & 3B & 45.03 & 25.26 & 35.15 & 84.20 & 33.07 & 16.05 & 24.56 & 76.85 & 37.13 & 35.62 & 36.38 & 80.96 \\
 & 11B   & 41.62 & 15.38 & 28.50 & 80.43 & 36.40 & 16.40 & 26.40 & 79.84 & 40.74 & 15.07 & 27.90 & 80.61 \\ \midrule
\multirow{4}{*}{Llama 1} & 7B   & 25.01 & 21.70 & 23.36 & 37.43 & 25.37 & 23.10 & 24.23 & 39.65 & 23.07 & 20.89 & 21.98 & 34.17 \\
 & 13B & 35.13 & 16.41 & 25.77 & 59.11 & 45.60 & 36.20 & 40.90 & 72.49 & 48.54 & 25.45 & 36.99 & 73.74 \\
 & 65B &  58.06 & 33.84 & 45.95 & 83.38 & 58.35 & 37.12 & 47.73 & 82.38 & 63.63 & 16.63 & 40.13 & 83.64 \\  \midrule
\multirow{3}{*}{Llama 2} & 7B  & 26.68 & 9.37 & 18.03 & 50.02 & 41.51 & 34.56 & 38.03 & 67.07 & 41.94 & 25.26 & 33.60 & 67.29 \\
 & 13B  & 62.02 & 35.69 & 48.86 & 83.08 & 56.12 & 45.05 & 50.58 & 82.05 & 58.55 & 31.89 & 45.22 & 83.65\\ \vspace{1.5mm} 
 & 70B  & 63.13 & 37.52 & 50.33 & 84.36 & 62.07 & 35.37 & 48.72 & 85.06 & 52.84 & 6.43 & 29.63 & 79.10\\ 
\multirow{3}{*}{Llama2chat} & 7B & 43.66 & 15.63 & 29.64 & 61.23 & 17.69 & 12.50 & 15.09 & 20.15 & 19.82 & 20.62 & 20.22 & 17.99\\
 & 13B   & 55.79 & 32.88 & 44.33 & 74.62 & 56.11 & 28.97 & 42.54 & 76.92 & 52.23 & 27.39 & 39.81 & 77.52   \\ 
 & 70B & 73.61 & 59.65 & 66.63 & 88.71 & 71.28 & 30.91 & 51.10 & 82.45 & 67.82 & - & 67.82 & 81.88 \\  \midrule
\multirow{2}{*}{Llama3} & 8B    & 64.17 & 48.52 & 56.35 & 86.50 & 57.43 & 35.72 & 46.58 & 85.85 & 37.86 & 9.51 & 23.69 & 78.80\\ \vspace{1.5mm} 
 & 70B  & 76.82 & 41.82 & 59.32 & 92.86 & 75.47 & 41.62 & 58.55 & 92.00 & 64.67 & 9.43 & 37.05 & 86.07  \\
\multirow{2}{*}{Llama3Instruct} & 8B & 53.08 & 29.74 & 41.41 & 88.86 & 50.43 & 13.25 & 31.84 & 85.51 & 37.34 & 3.80 & 20.57 & 79.64  \\
 & 70B  & 78.14 & 59.26 & 68.70 & 94.24 & 74.80 & 43.32 & 59.06 & 93.47 & 67.25 & 28.17 & 47.71 & 93.03 \\ \midrule \vspace{1.5mm} 
Mistral & 7B  & 56.79 & 26.70 & 41.75 & 84.15 & 55.94 & 37.50 & 46.72 & 84.87 & 51.19 & 13.06 & 32.13 & 83.21 \\ 
Mistral-Instruct & 7B  & 65.84 & 31.48 & 48.66 & 86.09 & 62.93 & 30.99 & 46.96 & 84.92 & 61.64 & 27.63 & 44.63 & 84.21 \\ \midrule
\multirow{2}{*}{Gemma} & 2B   & 30.18 & 26.26 & 28.22 & 37.77 & 26.93 & 27.99 & 27.46 & 45.90 & 28.41 & 22.11 & 25.26 & 47.42  \\ \vspace{1.5mm} 
 & 7B & 62.41 & 47.24 & 54.83 & 86.17 & 53.39 & 41.94 & 47.66 & 84.22 & 52.49 & 22.35 & 37.42 & 85.89\\
\multirow{2}{*}{Gemma-Instruct} & 2B  & 51.65 & 42.72 & 47.19 & 59.64 & 53.23 & 15.00 & 34.11 & 54.11 & 50.51 & 44.52 & 47.52 & 49.51\\ 
 & 7B  & 81.92 & 51.79 & 66.85 & 92.43 & 72.34 & 30.00 & 51.17 & 84.64 & 80.70 & 30.00 & 55.35 & 90.56 \\ \midrule 
Phi2 & 3B   & 30.09 & 18.27 & 24.18 & 54.92 & 37.21 & 19.02 & 28.11 & 67.34 & 38.48 & 15.78 & 27.13 & 68.16 \\ 
\bottomrule
\end{tabular}%
}
\caption{Consistency performance (Values are scaled by 100).}
\label{tab:consistency}
\end{table*}

\begin{table*}[h]
\centering
\resizebox{.7\textwidth}{!}{%
\begin{tabular}{cr|rr>{\columncolor[HTML]{E9EFF6}}r>{\columncolor[HTML]{E9EFF6}}r|rr>{\columncolor[HTML]{E9EFF6}}r>{\columncolor[HTML]{E9EFF6}}r|rr>{\columncolor[HTML]{E9EFF6}}r>{\columncolor[HTML]{E9EFF6}}r}
\toprule
\multirow{3}{*}{\textbf{Model}} & \multicolumn{1}{c}{\multirow{3}{*}{\textbf{Params}}} & \multicolumn{4}{c}{\textbf{zero-shot}} & \multicolumn{4}{c}{\textbf{four-shot}} & \multicolumn{4}{c}{\textbf{four-shot-2}} \\
\cmidrule(lr){3-6}
\cmidrule(lr){7-10}
\cmidrule(lr){11-14}
 & \multicolumn{1}{c}{} & \multicolumn{3}{c}{\textbf{Seen}} & \multicolumn{1}{c}{\textbf{Uneen}}  & \multicolumn{3}{c}{\textbf{Seen}} & \multicolumn{1}{c}{\textbf{Uneen}}  & \multicolumn{3}{c}{\textbf{Seen}} & \multicolumn{1}{c}{\textbf{Uneen}} \\
 \cmidrule(lr){3-5}
  \cmidrule(lr){6-6}
\cmidrule(lr){7-9}
\cmidrule(lr){10-10}
\cmidrule(lr){11-13}
\cmidrule(lr){14-14}
 & \multicolumn{1}{c}{} & \multicolumn{1}{c}{\textbf{CWR ($\downarrow$)}} & \multicolumn{1}{c}{\textbf{CCR ($\uparrow$)}} & \multicolumn{1}{c}{\cellcolor[HTML]{E9EFF6}\textbf{NCCR ($\uparrow$)}} & \multicolumn{1}{c}{\cellcolor[HTML]{E9EFF6}\textbf{IUR ($\uparrow$)}} 
 & \multicolumn{1}{c}{\textbf{CWR ($\downarrow$)}} & \multicolumn{1}{c}{\textbf{CCR ($\uparrow$)}} & \multicolumn{1}{c}{\cellcolor[HTML]{E9EFF6}\textbf{NCCR ($\uparrow$)}}  & \multicolumn{1}{c}{\cellcolor[HTML]{E9EFF6}\textbf{IUR ($\uparrow$)}} 
 & \multicolumn{1}{c}{\textbf{CWR ($\downarrow$)}} & \multicolumn{1}{c}{\textbf{CCR ($\uparrow$)}} & \multicolumn{1}{c}{\cellcolor[HTML]{E9EFF6}\textbf{NCCR ($\uparrow$)}}  & \multicolumn{1}{c}{\cellcolor[HTML]{E9EFF6}\textbf{IUR ($\uparrow$)}}  \\
 \toprule
  GPT-3.5 Turbo & -  & 18.78 & 52.90 & 34.11 & 95.67& 16.64 & 52.09 & 35.45 & 98.96& 13.56 & 45.66 & 32.09 & 99.79\\ \midrule
\multirow{5}{*}{Flan-T5} & 0.08B & 10.58 & 0.52 & -10.06 & 81.17& 14.13 & 0.77 & -13.36 & 75.09& 15.22 & 0.68 & -14.55 & 83.09 \\
 & 0.25B & 29.25 & 3.41 & -25.84 & 75.13& 28.83 & 3.66 & -25.17 & 75.25& 11.54 & 1.69 & -9.85 & 94.22 \\
 & 0.78B & 1.69 & 1.82 & 0.13 & 100.00& 12.91 & 6.01 & -6.90 & 95.04& 4.13 & 3.45 & -0.68 & 99.97 \\
 & 3B & 4.37 & 6.32 & 1.95 & 99.68& 15.21 & 10.51 & -4.70 & 96.19& 8.55 & 9.09 & 0.54 & 99.90 \\
 & 11B  & 17.05 & 16.70 & -0.35 & 95.01& 22.07 & 17.75 & -4.32 & 91.02& 17.26 & 16.28 & -0.98 & 98.49 \\ \midrule
\multirow{4}{*}{Llama 1} & 7B  & 10.99 & 13.35 & 2.36 & 83.38& 13.82 & 16.69 & 2.88 & 77.99& 6.36 & 9.66 & 3.30 & 90.50  \\
 & 13B & 12.06 & 24.31 & 12.25 & 87.63& 22.65 & 34.38 & 11.73 & 65.29& 13.28 & 26.20 & 12.91 & 92.20 \\
 & 65B & 16.49 & 39.08 & 22.59 & 78.94& 23.21 & 47.56 & 24.35 & 70.10& 9.31 & 29.16 & 19.85 & 98.35\\  \midrule
\multirow{3}{*}{Llama 2} & 7B  & 6.37 & 15.76 & 9.39 & 96.94& 22.68 & 28.19 & 5.51 & 67.28& 14.90 & 20.25 & 5.35 & 91.56\\
 & 13B  & 14.68 & 34.12 & 19.44 & 79.60& 27.05 & 40.34 & 13.29 & 57.67& 16.04 & 32.68 & 16.64 & 90.83\\ \vspace{1.5mm} 
 & 70B  & 23.47 & 46.68 & 23.21 & 69.30& 23.71 & 50.89 & 27.18 & 69.24& 10.30 & 36.41 & 26.11 & 99.69\\ 
\multirow{3}{*}{Llama2chat} & 7B & 20.68 & 22.24 & 1.56 & 93.80& 4.75 & 5.62 & 0.87 & 99.83& 4.64 & 4.20 & -0.45 & 99.94  \\
 & 13B & 21.13 & 30.80 & 9.67 & 90.56& 14.04 & 30.10 & 16.05 & 98.88& 16.71 & 31.99 & 15.28 & 98.60\\ 
 & 70B & 21.74 & 42.14 & 20.40 & 98.87& 10.39 & 28.12 & 17.73 & 99.89& 10.26 & 26.69 & 16.43 & 100.00\\  \midrule
\multirow{2}{*}{Llama3} & 8B  & 32.22 & 39.04 & 6.82 & 56.69& 28.30 & 41.41 & 13.11 & 66.60& 12.41 & 28.63 & 16.22 & 98.96 \\ \vspace{1.5mm} 
 & 70B & 21.41 & 51.57 & 30.15 & 71.62& 25.43 & 58.52 & 33.08 & 69.17& 12.20 & 46.13 & 33.93 & 98.83 \\
\multirow{2}{*}{Llama3Instruct} & 8B & 28.62 & 37.35 & 8.72 & 94.04& 27.23 & 33.58 & 6.34 & 95.95& 20.39 & 30.85 & 10.46 & 99.19\\
 & 70B  & 28.76 & 55.63 & 26.88 & 82.24& 28.72 & 54.31 & 25.59 & 86.34& 26.16 & 52.84 & 26.68 & 96.79 \\ \midrule \vspace{1.5mm} 
Mistral & 7B  & 13.63 & 33.21 & 19.58 & 86.15& 28.04 & 39.95 & 11.90 & 67.59& 12.65 & 30.26 & 17.62 & 97.61 \\ 
Mistral-Instruct & 7B & 29.48 & 25.74 & -3.74 & 92.60& 24.88 & 24.31 & -0.56 & 98.08& 28.74 & 24.82 & -3.92 & 94.24 \\ \midrule
\multirow{2}{*}{Gemma} & 2B  & 15.45 & 9.30 & -6.15 & 81.14& 18.62 & 12.42 & -6.20 & 72.78& 11.19 & 8.85 & -2.33 & 90.29\\ \vspace{1.5mm} 
 & 7B & 24.22 & 34.24 & 10.02 & 58.76& 30.16 & 34.14 & 3.97 & 61.70& 16.15 & 26.82 & 10.67 & 93.06 \\
\multirow{2}{*}{Gemma-Instruct} & 2B  & 27.79 & 5.53 & -22.26 & 95.13& 7.06 & 2.33 & -4.74 & 99.99& 7.27 & 1.87 & -5.41 & 99.69\\ 
 & 7B  & 30.42 & 17.59 & -12.83 & 99.27& 4.46 & 12.02 & 7.55 & 99.99& 15.44 & 12.32 & -3.12 & 99.98 \\ \midrule 
Phi2 & 3B  & 19.85 & 11.77 & -8.08 & 84.25& 26.83 & 14.41 & -12.42 & 83.79& 23.88 & 13.17 & -10.71 & 92.14  \\ 
\bottomrule
\end{tabular}%
}
\caption{Reliability performance (Values are scaled by 100).}
\label{tab:reliability}
\end{table*}

\label{sec:appendix}

\end{document}